\documentclass[sigconf, authorversion, nonacm]{acmart}
\usepackage[ruled,vlined,linesnumbered]{algorithm2e}
\usepackage{amsmath}
\usepackage{dsfont}
\usepackage{nicefrac}
\usepackage{graphicx}
\usepackage{subcaption}
\usepackage{dblfloatfix}

\AtBeginDocument{%
  }

\acmSubmissionID{1234}

\begin{document}

\title{Practical Batch Bayesian Sampling Algorithms for Online Adaptive Traffic Experimentation}

\author{Zezhong Zhang}
\email{zezzhang@ebay.com}
\affiliation{%
  \institution{eBay, Inc.} 
  \city{San Jose}
  \state{California}
  \country{USA}
}

\author{Ted Yuan}
\email{teyuan@ebay.com}
\affiliation{%
  \institution{eBay, Inc.} 
  \city{San Jose}
  \state{California}
  \country{USA}
}


\begin{abstract}
Online controlled experiments have emerged as industry gold standard for assessing new web features. 
As new web algorithms proliferate, experimentation platform faces an increasing demand on the velocity of online experiments, 
which encourages adaptive traffic testing methods to speed up identifying best variant by efficiently allocating traffic.
This paper proposed four Bayesian batch bandit algorithms (\textbf{NB-TS}, \textbf{WB-TS}, \textbf{NB-TTTS}, \textbf{WB-TTTS}) for eBay's experimentation platform, using summary batch statistics of a goal metric without incurring new engineering technical debts.
The novel \textbf{WB-TTTS}, in particular, demonstrates as an efficient, trustworthy and robust alternative to fixed horizon A/B testing.
Another novel contribution is to bring trustworthiness of best arm identification algorithms into evaluation criterion and highlight the existence of severe false positive inflation with equivalent best arms. 
To gain the trust of experimenters, experimentation platform must consider both efficiency and trustworthiness;
However, to the best of authors' knowledge, trustworthiness as an important topic is rarely discussed. 
This paper shows that Bayesian bandits without neutral posterior reshaping, particularly naive Thompson sampling (\textbf{NB-TS}), are untrustworthy because they can always identify an arm as the best from equivalent best arms.
To restore trustworthiness, a novel finding uncovers connections between convergence distribution of posterior optimal probabilities of equivalent best arms and neutral posterior reshaping, which controls false positives. 
Lastly, this paper presents lessons learned from eBay's experience, as well as thorough evaluation. 
We hope that this paper is useful to other industrial practitioners and inspires academic researchers interested in the trustworthiness of adaptive traffic experimentation.
\end{abstract}


\maketitle

\section{Introduction}
Online controlled experiments \cite{kohavi2017online} are critical in data-driven decision-making for online web businesses and widely accepted as the gold standard of statistical causal inference. 
The simplest and most common design is called A/B or A/B/n testing, where the traffic is allocated randomly and uniformly distributed among several arms (a.k.a. variations). 
With the statistical readout, decision-makers can identify and launch the winner confidently. 
Though large companies, like eBay, have large scale web traffic for online web experiments, live traffic is still limited valuable resource in this era of rapid iteration of new web algorithms (e.g., search, advertising, recommendation and risk algorithms).
Experimentation platforms are under increasing pressure to increase the velocity of online web experiments in order to identify potential excellent new web algorithms more quickly.
The number of new web algorithms in such a pilot experiment is commonly over five.
For example, recall and ranking, two components of a recommendation algorithm, each have three configurable hyperparameters. 
Experimenters must determine which of these nine arms is the best.
These arms may have minor differences with very significant business impact \cite{deng2013improving, xie2016improving}, which requires more traffic to identify. 
Further, more traffic divisions, due to more arms, require more traffic (hence need extended time) to ensure statistical power of identification, which is against rapid experimentation culture and agile development practice. 

Instead of A/B/n testing, adaptive traffic experimentation through multi-armed bandit algorithms, that adaptively determine the probability of being assigned to a variant, is recently rising as an essential complementary alternative. 
The most appealing Bayesian sampling technique is Thompson Sampling (TS) \cite{thompson1933}, which heuristically allocates traffic to each arm proportional to the probability of being the best arm based on Bayesian updating and aims to minimize the opportunity cost (a.k.a. regret) of showing inferior arms during the experiment. 
It performs well under practical batch settings with only batch statistics in data pipeline without new technical debts \cite{yahoo2011,facebook2019}.
Due to its simplicity and statistical property in regret, it has attracted a great deal of attention in Amazon\cite{amazon2017}, eBay\cite{ebay2020}, Google\cite{google2012,google2014}, Meta\cite{facebook2019}, Microsoft\cite{agrawal2012, agrawal2013}, Optimizely, Yahoo\cite{yahoo2011}, etc.
In this paper, we propose two Thompson Sampling based approaches: straightforward \textbf{NB-TS} and novel \textbf{WB-TS}. 
However, widely adopted Thompson Sampling is not optimal in terms of achieving the fastest possible rate of posterior convergence on the truth.
To allocate traffic more efficiently, we propose two Top-Two Thompson Sampling (TTTS) \cite{TTTS} based approaches for adaptive traffic experiments: \textbf{NB-TTTS} and \textbf{WB-TTTS}. 
Particularly, this paper promotes the novel algorithm \textbf{WB-TTTS} due to its robustness and better precision of identifying. 

Any testing methodologies must validate the trustworthiness instead of blindly chasing the efficiency. 
The most straightforward way is to run multiple AA tests with equivalent best arms and see if the proportion of false positives is excessively inflated. 
A severe untrustworthy case is that the method can almost surely identify an arm as the best from equivalent best arms, which breaks trust of experimenters.
To the best of the authors' knowledge, such validation of adaptive traffic experimentation methodologies, about which experimenters are most concerned, is rarely discussed in best arm identification or multi-armed bandit literatures. 
Beyond recall (a.k.a. power or sensitivity) and regret, this paper proposes false positive rate and precision as trustworthiness evaluation of best arm identification. 
Moreover, our novel finding shows that Bayesian bandits without neutral posterior reshaping, particularly naive Thompson sampling (\textbf{NB-TS}), exist severe false positive inflation. 
To control false positives, novel connections between convergence distribution of posterior optimal probabilities and neutral posterior reshaping are uncovered, which restores trustworthiness. 

Besides above trustworthiness validation, this paper presents evaluations with both real-world eBay experiments and reproducible synthetic datasets, which cover trustworthiness, recall, regret and robustness on non-stationary rewards.
Our evaluation demonstrates that TTTS based approaches, especially \textbf{WB-TTTS}, is a more promising alternative to fixed horizon A/B testing (i.e. uniform sampling) for pilot experiments.
The \textbf{WB-TTTS} exhibits competitive optimal recall that is significantly higher than uniform sampling, indicating top-tier efficiency in traffic allocation. 
It also outperforms on the precision of identifying and shows robustness against non-stationary reward trend. 
Of course, when experimenters have concerns of directing users to poor variants, either cost or ethical reasons, \textbf{NB-TS} or \textbf{WB-TS} should be considered for collecting samples and minimizing regret trials. 
Compared with aggressive \textbf{NB-TS}, \textbf{WB-TS} is robust and conservative with well controlled false positives. 

The rest of the paper is structured as follows. 
Section~\ref{section:problem_formulation} details problem formulation of Bayesian batch bandit algorithms for practical adaptive traffic experimentation; 
In section~\ref{section:sampling_rules}, we discuss the likelihood of Naive Batch \textbf{WB}, propose the likelihood of Weighted Batch \textbf{WB} to restore CLT and unbiasedness, and further derive two TS based approaches and two TTTS based approaches; 
Section~\ref{section:quality} reveals false positive inflation and restores trustworthiness;
Section~\ref{section:evaluation} presents eBay learnings with empirical evaluations;
We conclude this work in section~\ref{section:conclusions}.

\section{Problem Formulation}
\label{section:problem_formulation}
Under frequentist setting, supposing there exist total $K$ arms with unknown true means that denoted as $\boldsymbol{\theta^*} \equiv (\theta_1^*, \dots, \theta_K^*)$, we denote each arm by its index $k \in \{1, 2, \dots, K\}$ and possible alternative parameters vector by $\boldsymbol{\theta} \equiv (\theta_1, \dots, \theta_K)$. 
The goal of our experiment is to identify the best arm $I^*_1 = \arg\max_{1 \le i \le K} \theta_i^*$ with the highest mean among all $K$ arms in an adaptive traffic experiment.

Under batch setting that each batch can collect sample size $T_b$ in total, we denote each batch by its index $b \in \{1, 2, \dots, B\}$ with at most $B$ batches of samples due to traffic budget.
A Bayesian bandit algorithm can determine the traffic allocation $\boldsymbol{e}_{b} \equiv (e_{b,1}, \dots, e_{b,K})$ for the current batch $b$.
In the meantime, we can observe independent and identically distributed (IID) sample rewards in batch $b$.
We denote data collection $\mathcal{D}_b \equiv \{I_{b,i}, e_{b, I_{b,i}}, Y_{b,i}\}_{i=1}^{T_b}$ for the collected IID samples in batch $b$, where $I_{b,i}$ is index value of arm served sample $i$ in batch $b$ and $Y_{b,i}$ is the corresponding observed reward. 

\paragraph{Bayesian Batch Bandit}
By modifying the general procedure of Bayesian Thompson Sampling, we abstract the Bayesian batch bandit framework in Algorithm~\ref{algo:batch_bandit} to illustrate one nature extension in practice.
We denote $\pi_0(\boldsymbol{\theta})$ as the initial prior distribution and $L_b(\boldsymbol{\theta}) \equiv p(\mathcal{D} \mid \boldsymbol{\theta})$ as the likelihood function after batch $b$ data collected. 
The corresponding probability density function of posterior $\pi_b(\boldsymbol{\theta})$ can be derived considering data collected from batch $b$.

\begin{algorithm}
  \DontPrintSemicolon
  \SetAlgoLined
  \BlankLine
  
  $\mathcal{D} = \{\}$\;
  Initial $\pi_0(\boldsymbol{\theta})$ as prior distribution\;
  \For{$b=1,2,\dots,B$}{
    $\mathcal{D}_b=\{\}$\;
    Determine traffic allocation $\boldsymbol{e_b}$ based on $\pi_{b-1}(\boldsymbol{\theta})$\;
    \For{$i=1,2,\dots,T_b$}{
      Serve arm $I_{b,i} \sim \mathbf{Cat}(K, \boldsymbol{e_b})$\;
      Observe reward $Y_{b,i}$\;
      $\mathcal{D}_b = \mathcal{D}_b \cup (I_{b,i}, e_{b, I_{b,i}}, Y_{b,i})$\;
    }
    $\mathcal{D} = \mathcal{D} \cup \mathcal{D}_b$\;
    Bayesian updating $\pi_b(\boldsymbol{\theta}) \propto \pi_0(\boldsymbol{\theta})L_b(\boldsymbol{\theta} \mid \mathcal{D})$\;
  }

  \caption{Bayesian Batch Bandit Framework}
  \label{algo:batch_bandit}
\end{algorithm}

\paragraph{Batch Settings} 
There exist two types of settings in practice. 
One is fixed sample size batches, where each batch traffic size $T_b = \lambda$ is a deterministic constant. The other is fixed duration batches, where each batch traffic size is a stochastic value (e.g., $T_b \sim Pois(\lambda)$ can be seen as sampling from a Poisson distribution). 
In either case, batch traffic size shall have $\mathbb{E}(T_b) = \lambda$ and assume no adversarial behavior, which is a fairly weak assumption in practice. 

\paragraph{Batch Statistics} 
For the collected samples $\mathcal{D}_b$ in batch $b$, there exist several common summary batch statistics for a specific arm $k$, including batch sample count $n_{b,k} = \sum_{i=1}^{T_b}{\mathbb{I}[I_{b,i}==k]}$, sum of batch rewards $S_{b,k} = \sum_{i=1}^{T_b}{\mathbb{I}[I_{b,i}==k]Y_{b,i}}$, batch mean $\bar{Y}_{b,k} = \frac{S_{b,k}}{n_{b,k}}$ and sum of squared rewards $SS_{b,k} = \sum_{i=1}^{T_b}{(\mathbb{I}[I_{b,i}==k]Y_i)^{2}}$. 

\paragraph{Posterior Optimal Probability}
With the posterior $\pi_b(\boldsymbol{\theta})$ after batch $b$, we denote the optimal probability of arm $i$ as below.
\begin{equation}
  \label{eq:optimal_prob}
  \begin{aligned}
    \alpha_{b,i} &= \int \mathbb{I}[i==\arg\max_{1 \le k \le K}\theta_k]\pi_b(\boldsymbol{\theta})d\boldsymbol{\theta}
  \end{aligned}
\end{equation}
Note that only one arm can have an optimal probability over 50\%.

\paragraph{Traffic Allocation}
Based on above optimal probabilities, traffic allocation $\boldsymbol{\tilde{e}}_b$ for batch $b$ will be computed. 
A common way to avoid potential $0\%$ traffic for an arm in any batch is to keep minimal $\gamma \times 100\%$\footnote{Without loss of generality, this paper uses $\gamma=1\%$ as minimum traffic due to engineering consideration} traffic per arm. 
Hence, $\gamma K$ samples are uniformly sampled among $K$ arms, while the rest are allocated adaptively by computed probabilities $\boldsymbol{\tilde{e}}_b$. 
The hybrid sampling by Formula~\ref{eq:traffic} is,
\begin{equation}
  \label{eq:traffic}
  \begin{aligned}
    \boldsymbol{e}_b = \gamma + (1 - \gamma K) \boldsymbol{\tilde{e}}_b
  \end{aligned}
\end{equation}
where the $\boldsymbol{e}_b$ is the actual rollout traffic allocation for batch $b$.

\paragraph{Decision of Identification}
After $B$ batches of data collected (e.g., 1 week), experimenter can claim the winner with the largest posterior optimal probability $\alpha_{B,i}$ if $\max_{1 \le i \le K}{\alpha_{B,i}} > \delta$, where $\delta$ is a pre-determined decision threshold of identification (e.g., 90\%). 


\section{Sampling Rules for Batch Bandit}
\label{section:sampling_rules}
This section describes two types of the likelihood function $L_b(\boldsymbol{\theta} \mid \mathcal{D})$ of batch bandit for the Bayesian updating step in Algorithm~\ref{algo:batch_bandit}. 

\subsection{Naive Batch Likelihood}
Each arm's immediate rewards are independent identically distributed but sampling from certain unknown distribution.
Thankfully, the classical CLT can guarantee the asymptotical normality of proper normalized batch mean $\bar{Y}_{b,k}$ per arm with large enough batch sample size,
\begin{equation}
  \label{eq:classical_clt}
  \begin{aligned}
    \sqrt{n_{b,k}}(\bar{Y}_{b,k} - \theta^*_{k}) \xrightarrow{d} \mathcal{N}(0,\sigma_{k}^{2})
  \end{aligned}
\end{equation}
where rewards of each arm can be from any unknown distribution with mean $\theta^*_{k}$ and variance $\sigma_{k}^{2}$.
This is a fairly weak assumption since we commonly have large enough traffic to against the skewness of metric.

With convenient Gaussian likelihood per arm $\theta_{k}$, the conjugate Gaussian prior $\pi_0(\theta_{k})$ can be used and updated iteratively as
\begin{equation}
  \label{eq:naive_iterative_update}
  \begin{aligned}
    \pi_b(\theta_k) &\propto \pi_0(\theta_k) \prod_{i=1}^{b}{p(\sqrt{n_{b,k}}(\bar{Y}_{b,k} - \theta_{k}) \mid \theta_{k}; \sigma_{k})}
  \end{aligned}
\end{equation}
where likelihood $p(\cdot \mid \theta_{k}; \sigma_{k})$ is the probability density function of $\mathcal{N}(0,\sigma_{k}^{2})$ with known variance and $\pi_b(\theta_k) = \mathcal{N}(\mu_{b,k}, \frac{1}{\tau_{b,k}})$ is the updated posterior distribution (see Appendix \ref{appendix:naive_batch}). 

Formula~\ref{eq:naive_iterative_update} is a natural but questionable extension for Bayesian updating. 
The product of likelihoods in formula \ref{eq:naive_iterative_update} assumes that the observed batch statistics in the sequence are independent.
However, this is certainly not true. Due to the nature of any outcome-adaptive sampling, the sample size $n_{b,k}$ of an ongoing batch as a random variable depends on the historical observed batch statistics.
Hence, the batch statistics derived by Formula \ref{eq:classical_clt} are dependent on rewards and traffic in all previous batches.

\begin{figure*}[ht]
  \begin{subfigure}[b]{0.24\linewidth}
    \centering
    \includegraphics[width=\linewidth]{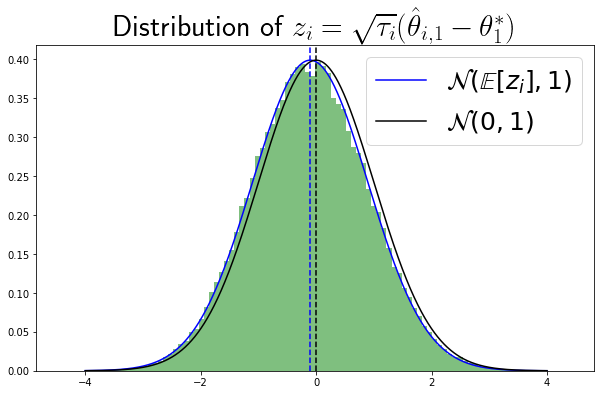} 
    \caption{$\hat{\theta}_{nb}$ By \textbf{NB-TS}} 
    \label{fig:unbiasedness:a} 
  \end{subfigure}
  \begin{subfigure}[b]{0.24\linewidth}
    \centering
    \includegraphics[width=\linewidth]{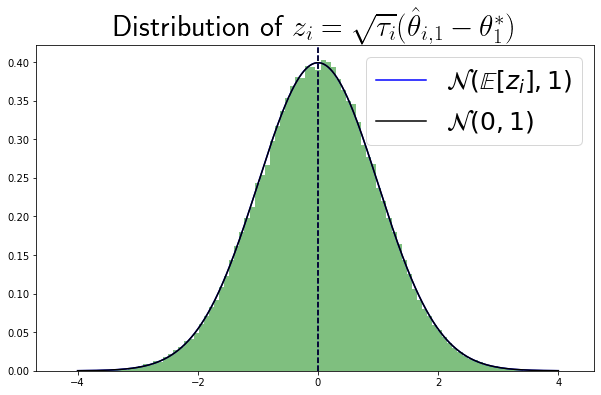} 
    \caption{$\hat{\theta}_{wb}$ By \textbf{WB-TS}}
    \label{fig:unbiasedness:b} 
  \end{subfigure}
  \begin{subfigure}[b]{0.24\linewidth}
    \centering
    \includegraphics[width=\linewidth]{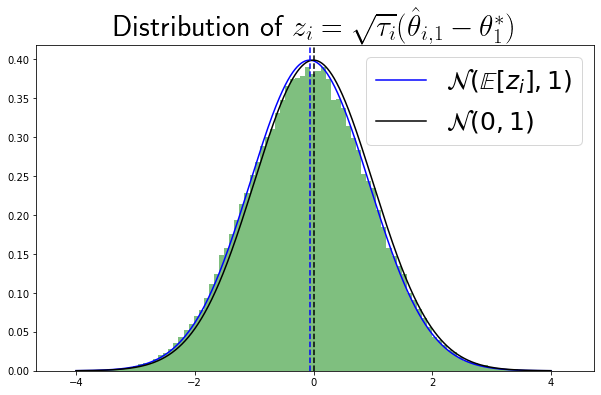} 
    \caption{$\hat{\theta}_{nb}$ By \textbf{NB-TTTS}} 
    \label{fig:unbiasedness:c} 
  \end{subfigure} 
  \begin{subfigure}[b]{0.24\linewidth}
    \centering
    \includegraphics[width=\linewidth]{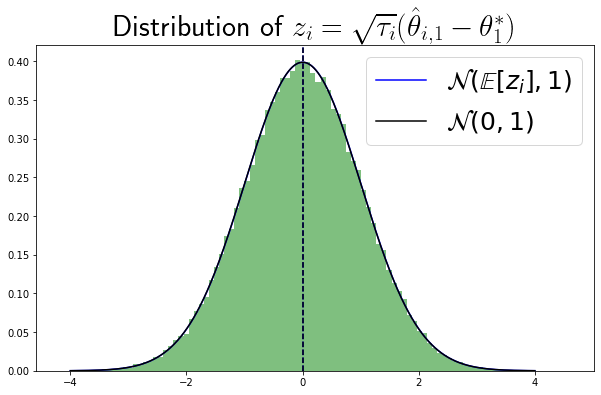} 
    \caption{$\hat{\theta}_{wb}$ By \textbf{WB-TTTS}} 
    \label{fig:unbiasedness:d} 
  \end{subfigure} 
  \caption{
    2-Batch 3-Arm Simulation: 
    We simulate 100,000 runs with two batches per run and 1,000 samples per batch.
    Three arms are uniformly sampled in the first batch, then adaptively sampled by four sampling rules in the second batch.
    The empirical distribution of $z_i = \sqrt{\tau_i} (\hat{\theta}_{i,1} - \theta^*_1)$ per run $i$ for the best arm are plotted.
  }
  \Description[2-Batch 3-Arm Simulation]{
    We simulate 100,000 runs with two batches per run and 1,000 samples per batch.
    Three arms are uniformly sampled in the first batch, then adaptively sampled by four sampling rules in the second batch.
    The empirical distribution of $z_i = \sqrt{\tau_i} (\hat{\theta}_{i,1} - \theta^*_1)$ for best arm per run $i$ are plotted.
    Note that $\sqrt{\tau_i}$ is the inverse standard error of estimation.
  }
  \label{fig:unbiasedness} 
  \vspace{-12pt}
\end{figure*}

By outcome-adaptive sampling, the naive batch statistics (Formula \ref{eq:nb_statistic}) by above likelihood are biased (see Figure~\ref{fig:unbiasedness:a} and \ref{fig:unbiasedness:c}),
\begin{equation}
  \label{eq:nb_statistic}
  \begin{aligned}
    \hat{\theta}_{nb}(k)=\frac{\sum_{i=1}^{b}{n_{i,k}\bar{Y}_{i,k}}}{\sum_{i=1}^{b}{n_{i,k}}} \not\xrightarrow{d} \mathcal{N}(\theta^*_{k}, \tau^{-1})
  \end{aligned}
\end{equation}
where inverse variance $\tau = \frac{1}{\sigma^2_{k}} \sum_{i=1}^{b}{n_{i,k}}$. 
And, such downward biased issue\footnote{Vitor Hadad et al. \cite{hadad2021confidence} pointed out that the inverse-probability weighting fixes the bias problem but results in a non-normal asymptotic distribution; especially, when assignment probability tends to zero, the tails of distribution becomes heavier.} is found and proved by many researchers \cite{hadad2021confidence, nie2018adaptively} when there exist no single best arm or a slightly better best arm. 
Hence, below proposes a better likelihood approximation. 

\subsection{Weighted Batch Likelihood}
Our idea is initially motivated by the weighted Z-test \cite{weighted_ztest}, which is a powerful meta-analysis method that combines multiple independent studies with weighting.
Although collected data batches under the outcome-adaptive sampling schema are not independent anymore, there may exist statistics with a proper weighting of these batches that can still satisfy asymptotical normality as,
\begin{equation}
  \label{eq:martingal_clt}
  \begin{aligned}
    \sum_{i=1}^{b}{\phi_{i,k}\sqrt{n_{i,k}}(\bar{Y}_{i,k} - \theta^*_{k})} \xrightarrow{d} \mathcal{N}(0,\sigma_{k}^{2}\sum_{i=1}^{b}{\phi_{i,k}^2})
  \end{aligned}
\end{equation}
where $\phi_{i,k}$ is the weight on batch $i$ for arm $k$. And, we denote the weights of batches for arm $k$ as $\boldsymbol{\phi}_k \equiv (\phi_{1,k}, \dots, \phi_{B,k})$. 

With proper weights, the weighted batch statistics with $b$ batches of data collected can be derived as below,
\begin{equation}
  \label{eq:wb_statistic}
  \begin{aligned}
    \hat{\theta}_{wb}(k) = \sum_{i=1}^{b}{w_{i,k} \bar{Y}_{i,k}} \xrightarrow{d} \mathcal{N}(\theta^*_{k}, \tau^{-1})
  \end{aligned}
\end{equation}
where $w_{i,k} = \frac{\phi_{i,k} \sqrt{n_{i,k}}}{\sum_{i=1}^{b}{\phi_{i,k} \sqrt{n_{i,k}}}}$ and variance $\tau^{-1} =  \sum_{i=1}^{b}{w_{i,k}^2 \frac{\sigma_{k}^{2}}{n_{i,k}}}$.

With independent batches, weighted Z-test \cite{weighted_ztest} suggested that the weights $\phi_{i,k}$ with minimal variance $\tau^{-1}$ can be achieved by $\phi_{i,k}=\sqrt{n_{i,k}}$, which is exactly the biased naive batch statistics by Formula \ref{eq:nb_statistic} under outcome-adaptive sampling.
Instead, we suggest $\phi_{i,k}=1$ or $\phi_{i,k}=\sqrt{T_i}$ as proper weights under outcome-adaptive sampling schema to restore the martingale CLT and adjust the bias. 
Such $\boldsymbol{\phi}$ with asymptotical normality property is called Weighted Batch (\textbf{WB}) in this paper, and can be claimed via variance stabilizing theorem\cite{hadad2021confidence} (see Appendix \ref{appendix:wb_statistics}).
Note such asymptotical normality property does not require asymptotical normality in each batch.

\paragraph{Choice of $\phi_{i,k}$}
On the choice between $\phi_{i,k}=1$ and $\phi_{i,k}=\sqrt{T_i}$, the two actually are the same under fixed sample size batch setting. 
While, under the fixed duration batch settings, each batch usually is differently powered by a different batch sample size.
Although $\phi_{i,k}=\sqrt{T_i}$ reflects the batch power by weighting, the power gain looks minor in practice. 
It is interesting to know whether there exists optimal power $\boldsymbol{\phi}$ with asymptotical normality property, but it is out of the scope of this paper.
In below discussion and simulation, we will use $\phi_{i,k} = 1$ without loss of generality. 

\paragraph{Unbiasedness}
To illustrate the unbiasedness property provided by $\hat{\theta}_{wb}$ (Formula~\ref{eq:wb_statistic}), let us run a simple 2-Batch 3-Arm simulation from $\mathcal{N}(\theta_k^*, 1)$, where $(\theta_1^*, \theta_2^*, \theta_3^*) = (0.01, 0, 0)$ with single best arm $k=1$.
The simulation settings are shown in Figure~\ref{fig:unbiasedness}. 
The empirical distribution of "studentized" $z_i = \sqrt{\tau_i} (\hat{\theta}_{i,1} - \theta^*_1)$ per run $i$ of the true best arm are plotted in green, and the corresponding Normal distribution by matching average value over runs are in blue curve.
By our simulations, it is evident that by $\hat{\theta}_{nb}$ Formula~\ref{eq:nb_statistic} are biased downward, even when there does exist a single best arm.
However, our proposed weighted batch statistics $\hat{\theta}_{wb}$ can be well approximated by Normal distribution with unbiasedness property. 

\subsection{WB-TS and WB-TTTS}
By applying the two types of likelihood function in Algorithm~\ref{algo:batch_bandit}, we have two schemas of weighting batches: \textbf{NB} and \textbf{WB}.
The two sampling rules \textbf{NB-TS} and \textbf{NB-TTTS} with \textbf{NB} are original TS and TTTS under batch setting.
Below propose two novel sampling algorithms: \textbf{WB-TS} and \textbf{WB-TTTS}.

\paragraph{Weighted Batch \textbf{WB}}
With proper weighted batch statistics in Formula~\ref{eq:wb_statistic}, the asymptotical and convenient Gaussian likelihood $L_b(\theta_{k} \mid \mathcal{D})$ can be easily constructed.
Then, we can leverage conjugate Gaussian prior $\pi_0(\theta_{k})=\mathcal{N}(\mu_{0,k}, \frac{1}{\tau_{0,k}})$\footnote{In this paper, we use improper prior by letting $\tau_{0,k}=0$ as an uninformative prior for each arm with $\pi_0(\theta_{k}) \propto 1$.} to derive the posterior distribution $\pi_b(\theta_k) = \mathcal{N}(\mu_{b,k}, \frac{1}{\tau_{b,k}})$ by Bayesian updating
\begin{equation}
\label{eq:weighted_update}
\begin{aligned}
  \tau_{b,k} &= \tau_{0,k} + \tau \\
  \mu_{b,k} &= \frac{\mu_{0,k}\tau_{0,k} + \tau \sum_{i=1}^{b}{w_{i,k} \bar{Y}_{i,k}} }{\tau_{0,k} + \tau}
\end{aligned}
\end{equation}
where $w_{i,k} = \frac{\phi_{i,k} \sqrt{n_{i,k}}}{\sum_{i=1}^{b}{\phi_{i,k} \sqrt{n_{i,k}}}}$ and $\tau = \frac{1}{\sigma_{k}^{2}} \frac{(\sum_{i=1}^{b}{\phi_{i,k}\sqrt{n_{i,k}}})^2}{\sum_{i=1}^{b}{\phi_{i,k}^2}} $.
For convenience, we name Algorithm~\ref{algo:batch_bandit} with Formula~\ref{eq:weighted_update} as \textbf{WB}.

Using proposed \textbf{WB} and $\tilde{e}_{b,k}=\alpha_{b,k}$ in Algorithm~\ref{algo:batch_bandit}, we have new sampling rules \textbf{WB-TS} to minimize the regret of choosing inferiors during an experiment.
However, it is not optimal on samples efficiency according to Daniel Russo's theoretical analysis\cite{TTTS}. 
Instead of Thompson Sampling, we promote a simple extension of Thompson Sampling called Top-Two Thompson Sampling (TTTS) to efficiently allocate traffic to identify best arm as early as possible, which appears to be missed in adaptive traffic experimentation.
Under frequentist setting, TTTS\cite{TTTS} is proved to achieve sharp optimal power of identifying if single best exists with 
\begin{equation}
  \label{eq:TTTS}
  \begin{aligned}
    \tilde{e}_{b,k} = \alpha_{b,k} \left(\beta + (1 - \beta) \sum_{i \neq k}{\frac{\alpha_{b,i}}{1 - \alpha_{b,i}}} \right)
  \end{aligned}
\end{equation}
where $\beta=\frac{1}{2}$ is used in this paper. 
With our proposed \textbf{WB}, the new sampling rule is called \textbf{WB-TTTS}.

\paragraph{Posterior Reshaping}
It is common to compute the posterior optimal probability $\boldsymbol{\alpha}_b$ from a reshaped distribution by introducing a reshaping parameter $\eta$ into the posterior distribution $\pi_b(\theta_k) = \mathcal{N}(\mu_{b,k}, \frac{1}{\eta \tau_{b,k}})$.
Generally, sharpening with $\eta > 1$ would favor exploitation, while widening with $\eta < 1$ would favor exploration. 

\paragraph{Sample Variance Estimation}
In practice, rewards variance $\sigma_k^2$ is unknown. It is straightforward to estimate sample variance by,
\begin{equation}
  \label{eq:sample_variance}
  \begin{aligned}
    \hat{\sigma}_k^2 = \frac{\sum_{i=1}^{b}{SS_{i,k}}}{\sum_{i=1}^{b}{n_{i,k}}} - \hat{\theta}(k)^2
  \end{aligned}
\end{equation}
where $\hat{\theta}(k)$ is the mean estimator using either $\hat{\theta}_{wb}(k)$ from Formula~\ref{eq:wb_statistic} or $\hat{\theta}_{nb}(k)$ from Formula~\ref{eq:nb_statistic}.
For ratio metric $\frac{\bar{X}_{i,k}}{\bar{Y}_{i,k}}$ per batch $i$, it is required to apply delta method \cite{deng2018applying,nie2020dealing}.

\section{Trustworthiness Concern and Cure}
\label{section:quality}
The testing procedure can be seen as making decision among multiple hypotheses as below.
\begin{equation}
  \label{eq:hypothesis}
  \begin{aligned}
    H_0 &\colon \text{$K^{\prime}$ arms are equivalent best} \\
    H_k &\colon I_{1}^{*} = k
  \end{aligned}
\end{equation}
where $H_0$ is the null hypothesis and there exist total $K$ alternative hypotheses.
And, each alternative hypothesis $H_k$ states that the arm $k$ is the best. 
Without loss of generality, we let arm $k=1$ as the best arm to be identified ($H_1$ is true) for situation that $H_0$ is false and true means of arms are in the decreasing order $\theta_1^* \ge \theta_2^* \ge \dots \ge \theta_K^*$.

This section presents our novel findings on false positive inflation when null hypothesis $H_0$ is true, which is the trustworthiness concern from experimenters under frequentist settings.
And, a novel neutral posterior reshaping is proposed to restore trustworthiness.

\subsection{False Positive Rate Inflation}

\begin{figure*}[ht]
  \begin{subfigure}[b]{0.24\linewidth}
    \centering
    \includegraphics[width=\linewidth]{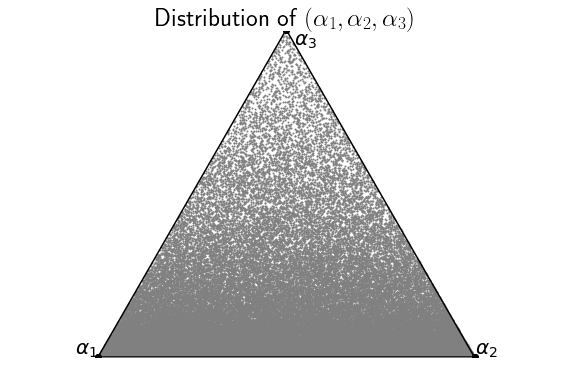} 
    \caption{$\boldsymbol{\alpha}$ of First Batch} 
    \label{fig:converge_case:a} 
  \end{subfigure} 
  \begin{subfigure}[b]{0.24\linewidth}
    \centering
    \includegraphics[width=\linewidth]{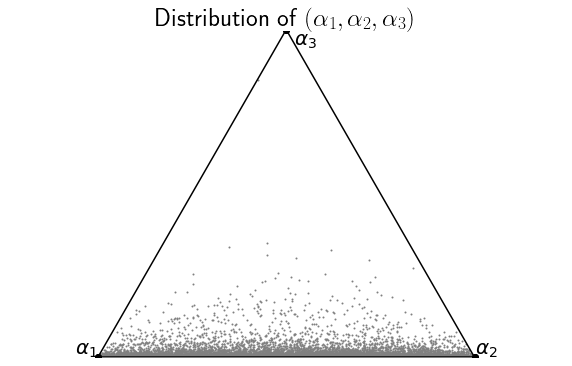} 
    \caption{$\boldsymbol{\alpha}$ of Last Batch (except \textbf{NB-TS})} 
    \label{fig:converge_case:b} 
  \end{subfigure} 
  \begin{subfigure}[b]{0.24\linewidth}
    \centering
    \includegraphics[width=\linewidth]{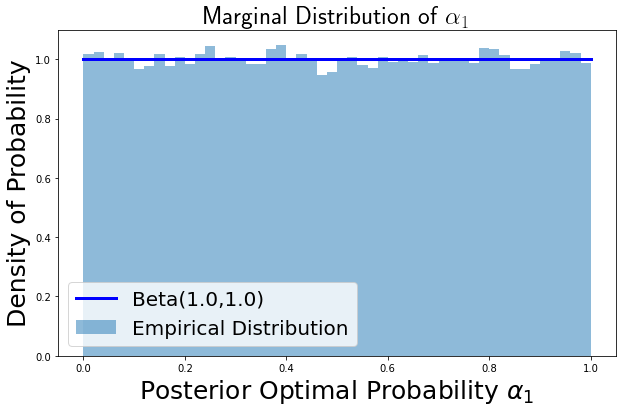} 
    \caption{$\alpha_1$ of Last Batch (except \textbf{NB-TS})} 
    \label{fig:converge_case:c} 
  \end{subfigure}
  \begin{subfigure}[b]{0.24\linewidth}
    \centering
    \includegraphics[width=\linewidth]{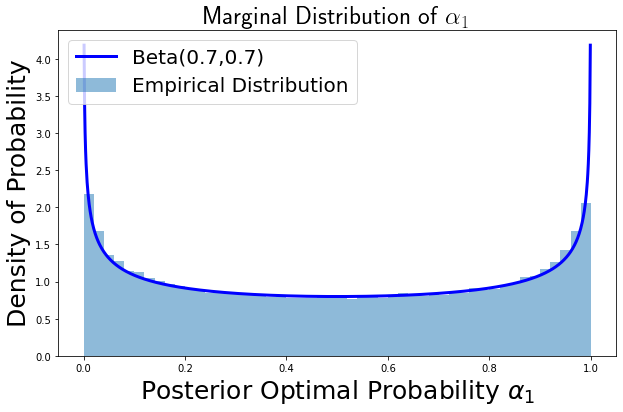} 
    \caption{$\alpha_1$ of Last Batch (by \textbf{NB-TS})} 
    \label{fig:converge_case:d} 
  \end{subfigure} 
  \caption{
    Simulation with $\boldsymbol{\theta}^*=(0.25,0.25,0)$ and unit variance: 
    We simulate 100,000 runs with 30 batches per run and 300 samples per batch per arm with $\eta=1$.
    The inferior arm $\alpha_3$ converges to almost zero and equivalent best arms $\alpha_{K^\prime}=(\alpha_1, \alpha_2)$ approximately converges to likely uniform.
    Note that the blue Beta curves are fitted by matching moments.
  }
  \Description[$K=3$ and $K^{\prime}=2$ Simulation]{
    We simulate 100,000 runs with 30 batches per run and 30 samples per batch per arm. The three arms are uniformly sampled in each batch.
  }
  \label{fig:converge_case}
  \vspace{-12pt}
\end{figure*}

Any experienced experimenters should leverage "AA tests" to explore if there exist any trustworthiness concerns for new testing methodologies. 
However, Trustworthiness of Bayesian bandits based adaptive traffic experiments are seldom evaluated.
In eBay, many experimenters, including us, conducted real experiments with equivalent best arms $K=K^{\prime}=2,3$. 
We found that Thompson sampling (or other default Bayesian bandits) would pick a winner much more often than A/B testing for situations with equivalent best arms. 
The inflated false positive rate (FPR) raised up trustworthiness concerns, however, there exist very few literatures on this topic, which motivated us to share our findings in front of broader audience. 
Below presents evaluations with datasets $A$ ($K^{\prime}=2$), $B$ ($K^{\prime}=3$), $C$ (real experiments with probably both $K^{\prime}=2,3$) and simplified simulations to reproduce and demonstrate false positive inflation found in eBay practice.

In Table~\ref{tab:fpr_inflation}, when there exist more than 2 equivalent best, except our proposed \textbf{WB-TS}, all others with default posterior reshaping $\eta=1$ causes FPR inflation. 
Especially, \textbf{NB-TS} with $\eta=1$ shows obvious inflation against both situations of equivalent best arms. 

\begin{table}[ht]
  \centering
  \begin{tabular}{|c|c|c|c|c|c|}
  \hline
                 & Unif            & NB-TS           & WB-TS  & NB-TTTS         & WB-TTTS         \\ \hline
  A ($\eta=1$)   & 8.8\%           & \textbf{14.3\%} & 7.4\%  & 9.6\%           & 8.8\%           \\ \hline
  B ($\eta=1$)   & \textbf{16.9\%} & \textbf{17.9\%} & 9.7\%  & \textbf{18.1\%} & \textbf{14.2\%} \\ \hline
  B ($\eta=0.7$) & 8.8\%           & 7.5\%           & 3.5\%  & 8.9\%           & 6.6\%           \\ \hline
  C ($\eta=1$)   & 12.8\%          & \textbf{26.5\%} & 10.4\% & \textbf{17.5\%} & 12.6\%          \\ \hline
  C ($\eta=0.7$) & 6.9\%           & 8.7\%           & 5.1\%  & 7.4\%           & 6.3\%           \\ \hline
  \end{tabular}
  \caption{False Positive Rate By Nominal FPR $\rho=10\%$}
  \label{tab:fpr_inflation}
  \vspace{-12pt}
\end{table}

Moreover, in Figure~\ref{fig:converge_case}, our simulations on $K=3$ and $K^{\prime}=2$ shows that the $\boldsymbol{\alpha}_{K^{\prime}=2}$ approximately converges to $\mathbf{Beta}(1,1)$, except Thompson sampling (\textbf{NB-TS}). 
\textbf{NB-TS} converges around corners of the two equivalent variants, causing severe false positive inflation.

\begin{figure}[ht]
  \begin{subfigure}[b]{0.5\linewidth}
    \centering
    \includegraphics[width=\linewidth]{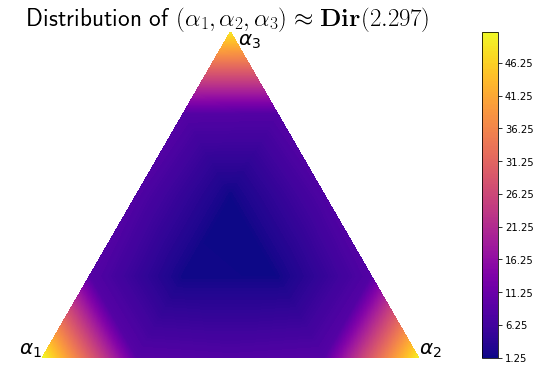} 
    \caption{Empirical $\boldsymbol{\alpha}_{K^{\prime}}$ ($\eta=1$)} 
    \label{fig:3-eq-arms:a} 
  \end{subfigure}
  \begin{subfigure}[b]{0.5\linewidth}
    \centering
    \includegraphics[width=\linewidth]{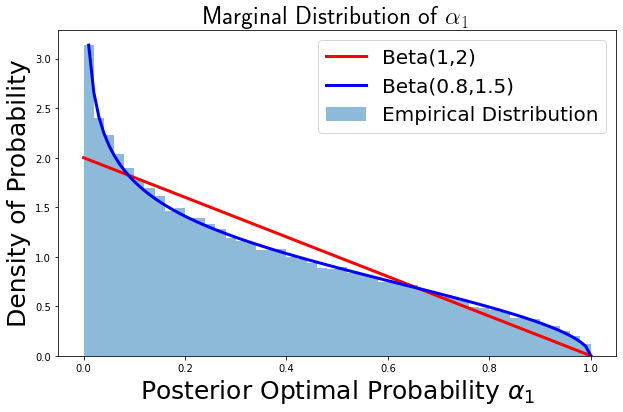} 
    \caption{Marginal $\alpha_1$ ($\eta=1$)}
    \label{fig:3-eq-arms:b} 
  \end{subfigure}
  \vskip\baselineskip
  \begin{subfigure}[b]{0.5\linewidth}
    \centering
    \includegraphics[width=\linewidth]{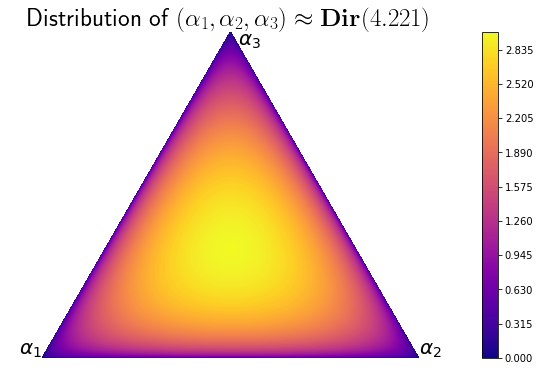} 
    \caption{Empirical $\boldsymbol{\alpha}_{K^{\prime}}$ ($\eta=0.5$)} 
    \label{fig:3-eq-arms:c} 
  \end{subfigure}
  \begin{subfigure}[b]{0.5\linewidth}
    \centering
    \includegraphics[width=\linewidth]{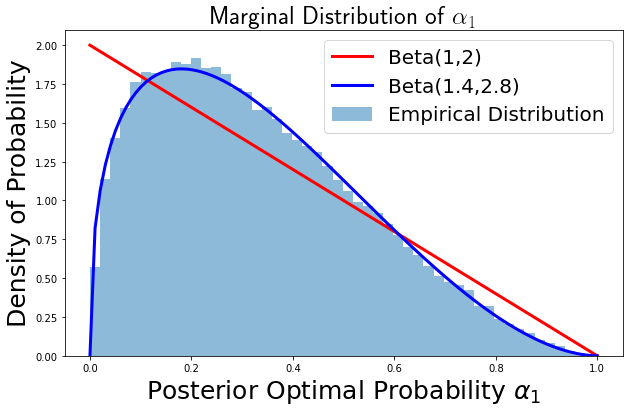} 
    \caption{Marginal $\alpha_1$ ($\eta=0.5$)}
    \label{fig:3-eq-arms:d} 
  \end{subfigure}
  \vskip\baselineskip
  \begin{subfigure}[b]{0.5\linewidth}
    \centering
    \includegraphics[width=\linewidth]{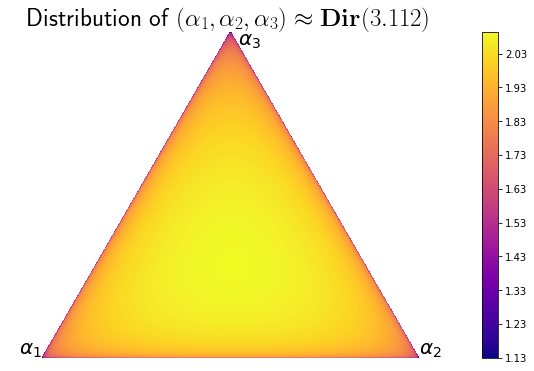} 
    \caption{Empirical $\boldsymbol{\alpha}_{K^{\prime}}$ ($\eta=0.7$)} 
    \label{fig:3-eq-arms:e} 
  \end{subfigure}
  \begin{subfigure}[b]{0.5\linewidth}
    \centering
    \includegraphics[width=\linewidth]{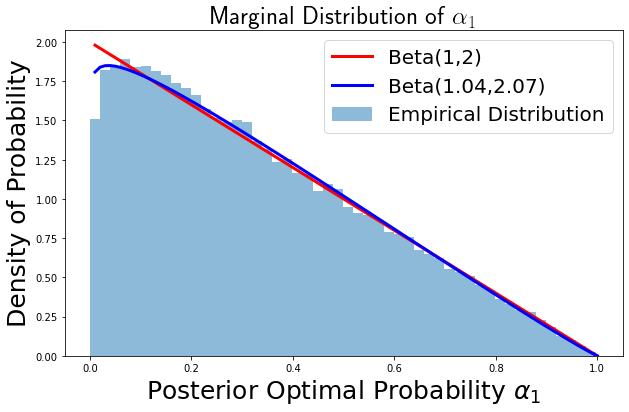} 
    \caption{Marginal $\alpha_1$ ($\eta=0.7$)}
    \label{fig:3-eq-arms:f} 
  \end{subfigure}
  \caption{
    To approximately restore flat Dirichlet with proper $\eta$, we simulate 100,000 runs with 10,000 samples per arm per run for $K=K^{\prime}=3$ situation with different $\eta$ by uniform sampling.
    In Figure~\ref{fig:3-eq-arms:a}-\ref{fig:3-eq-arms:f}, $K^{\prime}=3$ situation are reported with $\eta=1$, $\eta=0.5$ and grid search best $\eta=0.7$.
    Note that the red Beta curves are the theoretical marginal distribution $\mathbf{Beta}(1, K^{\prime}-1)$ of corresponding flat Dirichlet $\mathbf{Dir}(K^{\prime})$.
  }
  \Description[convergence of $K^{\prime}=3$]{
    convergence of $K^{\prime}=3$
  }
  \label{fig:3-eq-arms} 
  \vspace{-12pt}
\end{figure}

Further, more simulation evidences with $K=K^{\prime}=3$ show that the converged $\boldsymbol{\alpha}_{K^{\prime}=3}$ is far from $\mathbf{Dir}(3)$ even with uniform sampling.
In Figure~\ref{fig:3-eq-arms:a}, empirical distribution of $(\alpha_1,\alpha_2,\alpha_3)$ with $\eta=1$ are sparse and the vast majority of mass will be concentrated at the corners. That is to say, if we run multiple $K^{\prime}$ experiments with $\eta=1$ or larger, naive Bayesian bandit algorithms will eventually almost surely claim a single winner among the $K^{\prime}$ arms. 

\subsection{Theoretical Analysis: Control False Positives}
Daniel Russo\cite{TTTS} shows that posterior optimal probability of inferior arms almost surely converges to 0 as more evidence is collected under any sampling rules.
Intuitively, the optimal probability of rest $K^{\prime}$ arms (denote as $\boldsymbol{\alpha}_{K^{\prime}}$) can not converge to a single point but to a particular symmetric distribution (see Figure~\ref{fig:converge_case}). 
Mathematically, after convergence of optimal probabilities of inferior arms, the optimal probability $\boldsymbol{\alpha}_{K^{\prime}}$ of the rest $K^{\prime}$ equivalent best arms can be seen as symmetric $K^{\prime}$-dimensional Dirichlet distribution.
Especially, we denote $\boldsymbol{\alpha}_{K^{\prime}} \sim \mathbf{Dir}(a_0)$\footnote{The concentration parameter $a_0$ is defined as the sum of Dirichlet parameters of dimensions and $\boldsymbol{\alpha}_{K^{\prime}}$ is uniform distributed over $(K^{\prime}-1)$-simplex if $a_0=K^{\prime}$.}.

Motivated by the uniform nature of p-values under $H_0$ is true, we can control false positives by setting proper threshold $\delta$ on optimal probability if $\boldsymbol{\alpha}_{K^{\prime}}$ are uniformly distributed over the standard $(K^{\prime}-1)$-simplex.
In other words, we expect that the $\boldsymbol{\alpha}_{K^{\prime}}$ can converge to a flat $K^{\prime}$-dimensional Dirichlet distribution.

\begin{figure}[ht] 
  \begin{subfigure}[b]{0.45\linewidth}
    \centering
    \includegraphics[width=\linewidth]{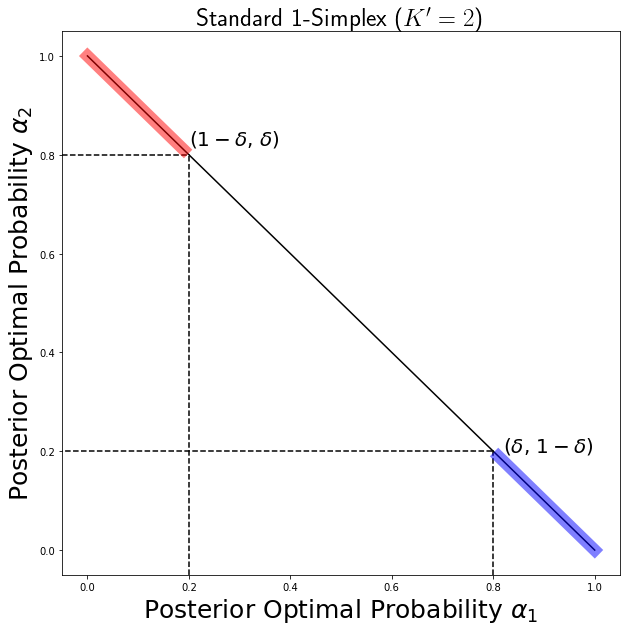} 
    \caption{$K^{\prime} = 2$} 
    \label{fig:simplex:a} 
  \end{subfigure}
  \begin{subfigure}[b]{0.55\linewidth}
    \centering
    \includegraphics[width=\linewidth]{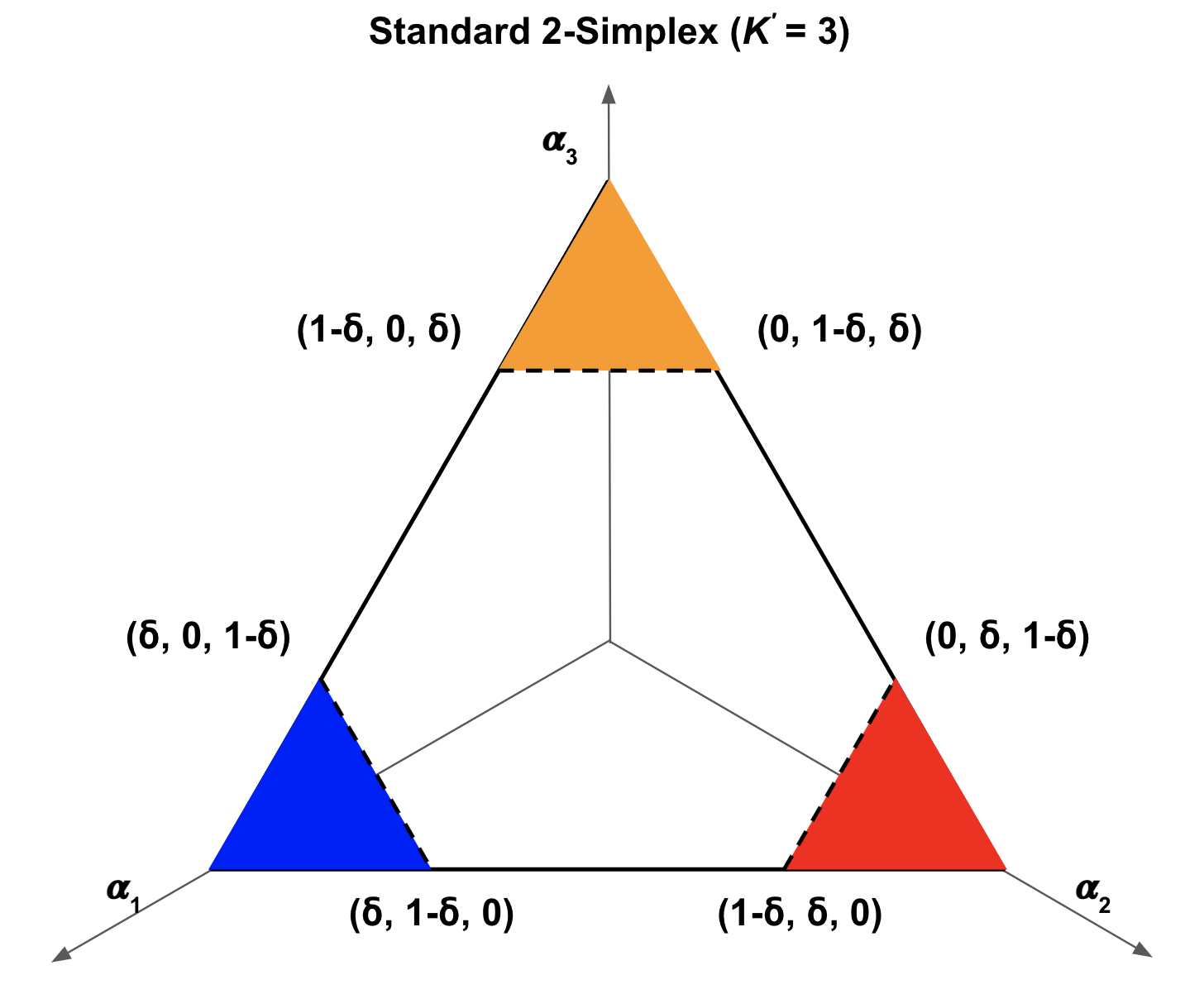} 
    \caption{$K^{\prime} = 3$}
    \label{fig:simplex:b} 
  \end{subfigure}
  \caption{
    If $\boldsymbol{\alpha}_{K^{\prime}}$ are uniformly distributed over the standard $(K^{\prime}-1)$-simplex, 
    by threshold $\delta$, the probability of making a Type I Error is the proportion of the colored area.
    It can be easily proved that the integration of area $\alpha_{1} > \delta$ (blue area) is $A_{K^{\prime}}(\delta) = (1-\delta)^{K^{\prime}-1} \frac{\sqrt{K^{\prime}}}{(K^{\prime}-1)!}$ for any $(K^{\prime}-1)$-simplex.
  }
  \Description[Standard $(K^{\prime}-1)$-Simplex]{
    The $\boldsymbol{\alpha}_{K^{\prime}}$ are uniformly distributed over the standard $(K^{\prime}-1)$-simplex. 
    By setting threshold $\delta$, the probability of a making Type I Error is the proportion of the colored area.
  }
  \label{fig:simplex} 
  \vspace{-12pt}
\end{figure}

When $H_0$ is true, suppose the $\boldsymbol{\alpha}_{K^{\prime}}$ can converge to a symmetric and flat $K^{\prime}$-dimensional Dirichlet distribution $\mathbf{Dir}(K^{\prime})$, the probability of making Type I Error \footnote{The probability of making Type I Error is denoted as $\rho$ in this paper, as notation $\alpha$ commonly represents posterior optimal probability in Bayesian bandit algorithms.} with infinite evidence will be
\begin{equation}
  \label{eq:type_i_error}
  \begin{aligned}
    \rho &= \mathbf{Pr}(\max{\boldsymbol{\alpha}_{K^{\prime}}} > \delta \mid H_0) \\
    &= K^{\prime} \mathbf{Pr}(\alpha_{1} > \delta \mid H_0) \\
    &= K^{\prime} (1 - \delta)^{K^{\prime} - 1}
  \end{aligned}
\end{equation}
Due to the $K^{\prime}$ arms are symmetric, we only need to figure out the $\mathbf{Pr}(\alpha_{1} > \delta \mid H_0)$, where $\alpha_{1}$ is observed optimal probability of the first arm in the $K^{\prime}$ arms. 
If $\boldsymbol{\alpha}_{K^{\prime}}$ are uniformly distributed over the standard $(K^{\prime}-1)$-simplex, $\mathbf{Pr}(\alpha_{1} > \delta \mid H_0)$ will be the proportion of simplex area with $\alpha_{1} > \delta$ over total standard $(K^{\prime}-1)$-simplex area (see Figure~\ref{fig:simplex} for understanding). 

To control false positive rate with pre-determined $K^{\prime}$ equivalent best arms, the proper threshold of optimal probability will be
\begin{equation}
  \label{eq:proper_delta}
  \begin{aligned}
    \delta = 1 - \left( \frac{\rho}{K^{\prime}} \right)^{\frac{1}{K^{\prime}-1}}
  \end{aligned}
\end{equation}
where $\rho$ is the probability of making Type I Error to be controlled\footnote{Specifically, this paper uses $\rho = 10\%$ without loss of generality.}.


\subsection{Neutral Posterior Reshaping}
To control false positives, this paper proposes the neutral posterior reshaping, which is to apply neutral $\eta$ that can restore flat Dirichlet $\mathbf{Dir}(K^{\prime})$.
In Figure~\ref{fig:3-eq-arms}, we can conclude that if we run with either larger or smaller $\eta$, the empirical distribution would be concentrated at either the corners or the center.
By grid search simulations, a neutral $\eta$ is proved to restore flat Dirichlet $\mathbf{Dir}(K^{\prime})$ in Figure~\ref{fig:3-eq-arms:e} and theoretically can control false positives. 

In Figure~\ref{fig:3-eq-arms:a}-\ref{fig:3-eq-arms:f}, three simulations on $K^{\prime}=3$ are reported.
The approximately neutral posterior reshaping $\eta=0.7$ is found by grid search, in order to match the moments of theoretical marginal distribution $\mathbf{Beta}(1, K^{\prime}-1)$.
Such neutral posterior reshaping $\eta=0.7$ allows the convergence of $\boldsymbol{\alpha}_{K^{\prime}=3}$ to be approximately flat Dirichlet distribution $\mathbf{Dir}(3)$. 
By grid search, more neutral $\eta$ are reported for $K^{\prime}=4$ ($\eta=0.57$) and $K^{\prime}=5$ ($\eta=0.5$).

\begin{figure}[ht]
  \centering
  \includegraphics[width=0.8\linewidth]{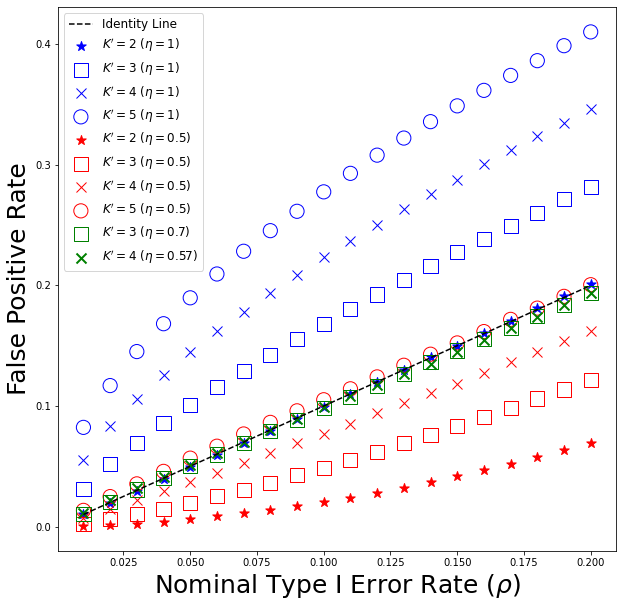}  
  \caption{
    False Positive Rate with $K=K^{\prime}=2,3,4,5$
  }
  \Description[Validation on $K^{\prime}=2,3$]{
    Validation on $K^{\prime}=2,3$
  }
  \label{fig:eta-rho} 
  \vspace{-6pt}
\end{figure}

Further, to validate the performance of controlling false positive rate (FPR), in Figure~\ref{fig:eta-rho}, FPR from these simulations are also reported. 
The FPR with $\eta=1$ (blue shapes) are inflated except for $K^{\prime}=2$ case, as $\eta=1$ is the neutral $\eta$ for $K^{\prime}=2$ case. 
If applying conservative $\eta=0.5$, FPR (red shapes) are deflated except for $K^{\prime}=5$ case. 
These neutral $\eta$ perform very well on controlling FPR around corresponding nominal $\rho$.

In conclusion, above neutral posterior reshaping\footnote{In the rest of paper, we will only consider $H_0$ cases with $K^{\prime}=2$ (using $\eta=1$) and $K^{\prime}=3$ (using $\eta=0.7$) since they are seemly more common situations in real-world. } restores flat Dirichlet and controls false positive rate.
Next section presents more FPR control learnings by applying such neutral posterior reshaping. 

\section{More Evaluations and Learnings}
\label{section:evaluation}
This section further evaluates the four Bayesian sampling rules and reveals learnings using synthetic and real-world experiments. 
This paper discusses robustness and evaluates FPR (False Positive Rate), Power (a.k.a. Recall), Precision and Regret (see Appendix~\ref{appendix:quality}).

\subsection{Synthetic and Real Experiments}
\label{subsection:datasets} 
This paper builds 4 synthetic datasets (20 stationary and 20 non-stationary experiments) and 20 real-world experiments for simulation and replay (see Table~\ref{tab:datasets} and Appendix~\ref{appendix:synthetic} for detail settings). 

\begin{table}[h]
  \centering
  \begin{tabular}{|c|c|c|c|c|c|}
  \hline
  Datasets     & $K$    & $K^{\prime}$ & Stationary & Noise            & \# of Tests \\ \hline
  A            & 10     & 2            & Y          & Gaussian         & 10          \\ \hline
  $A^{\prime}$ & 10     & 2            & N          & Gaussian         & 10          \\ \hline
  B            & 10     & 3            & Y          & Gaussian         & 10          \\ \hline
  $B^{\prime}$ & 10     & 3            & N          & Gaussian         & 10          \\ \hline
  C            & 7      & 2 or 3       & N          & Real Tests       & 10          \\ \hline
  \end{tabular}
  \caption{Datasets for Simulation and Replay}
  \label{tab:datasets}
  \vspace{-12pt}
\end{table}

Each synthetic dataset contains 10 experiments with $K=10$ arms, where half have a single best arm ($H_1$ is true) and half have $K^{\prime}$ equivalent best arms ($H_0$ is true).
The expected rewards in $A^{\prime}$ and $B^{\prime}$ are non-stationary over time. The difference of expected rewards between any of the two arms does not change, hence there still can exist a single best arm to be identified. 
The initial expected rewards of each arm in $A^{\prime}$ and $B^{\prime}$ are the same as the ones of corresponding stationary datasets ($A$ and $B$). 
To evaluate with the 40 synthetic tests, we simulate 100,000 runs with 20 batches per run and 500 samples per batch in each experiment. 

The dataset $C$ contains 10 real eBay cases with $K=7$ arms: 5 experiments with a single best arm and 5 experiments with unknown (probably $K^{\prime}=2,3$) equivalent best arms, according to the readout from original AB experiments.  
Samples of each experiment are collected from a metric of real A/B test as ground truth and selected one week traffic, where each has over ten million samples. 
We can observe probably single best arm if it has statistical significance compared with others in real A/B test, otherwise, it probably has equivalent best arms. 
To evaluate, we replay each test 100,000 runs with average 1 million samples per day. 
The replay rewards are randomly sampled with replacement from the corresponding day and arm of collected samples in the real A/B tests. 

\subsection{Results and eBay Learnings}
\label{subsection:perf}

\begin{figure}[ht]
  \begin{subfigure}[b]{0.5\linewidth}
    \centering
    \includegraphics[width=\linewidth]{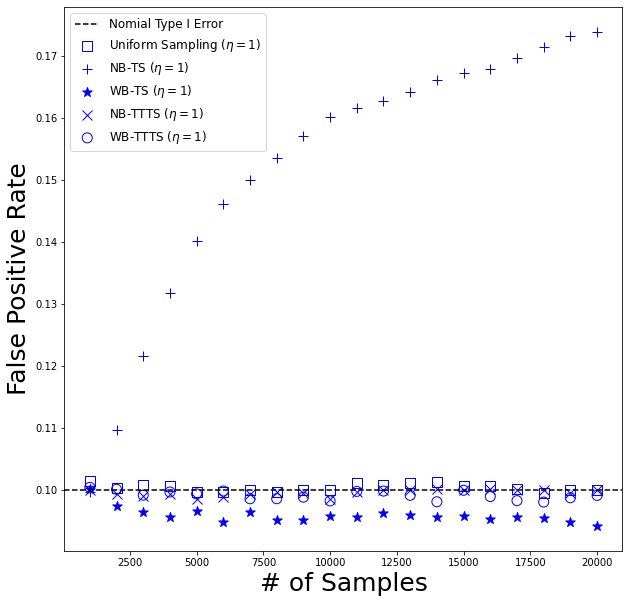} 
    \caption{$K=K^{\prime}=2$} 
    \label{fig:eta-fpr:a} 
  \end{subfigure}
  \begin{subfigure}[b]{0.5\linewidth}
    \centering
    \includegraphics[width=\linewidth]{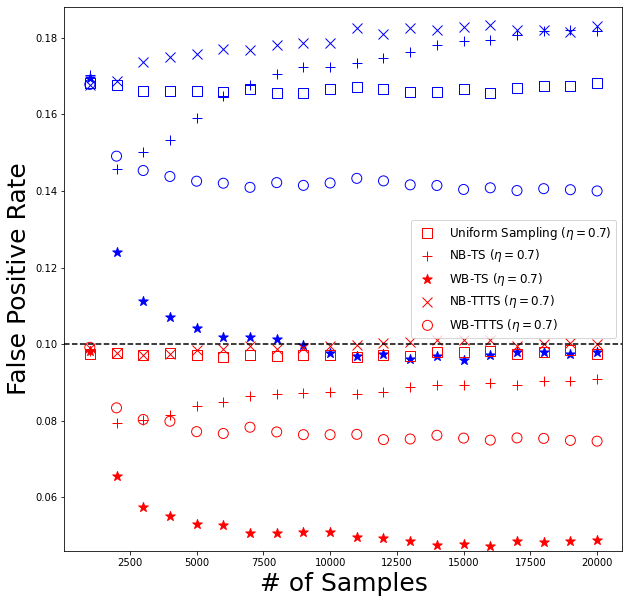} 
    \caption{$K=K^{\prime}=3$}
    \label{fig:eta-fpr:b} 
  \end{subfigure}
  \caption{
    To validate FPR control, experiments with $K=K^{\prime}=2,3$ are simulated with 100,000 runs and 1,000 samples per batch per run for each experiment.
  }
  \Description[Validation on $K^{\prime}=2,3$]{
    Validation on $K^{\prime}=2,3$
  }
  \label{fig:eta-fpr} 
  \vspace{-12pt}
\end{figure}

\paragraph{Both $\eta$ and \textbf{WB}-statistics help control false positives}
In Table~\ref{tab:fpr_inflation}, we also present FPR with $\eta=0.7$ on datasets $A$ ($K^{\prime}=2$), $B$ ($K^{\prime}=3$) and real experiments $C$ (probably $K^{\prime}=2,3$). 
It demonstrates that neutral $\eta$ can successfully control FPR under nominal $\rho=10\%$, especially for $K^{\prime}=3$ cases. 
More reproducible evidences with $K=K^{\prime}=2,3$ settings are shown in Figures~\ref{fig:eta-fpr:a}-\ref{fig:eta-fpr:b}. 
Clearly, neutral posterior reshaping theoretically works well and rule of thumb $\eta=1, 0.7$ does significantly reduce false positives.
However, there exist exceptions with \textbf{NB-TS} ($\eta=1$), causing severe inflation. 
While, interestingly, \textbf{WB-TS} ($\eta=1$) can control FPR with default posterior reshaping for both $K^{\prime}=2, 3$ cases.
It seems that the default $\eta=1$ is already a neutral posterior reshaping for \textbf{WB-TS} sampling rule. 
Such probably is one of the benefits from weighted batch likelihood (\textbf{WB}) in Formula~\ref{eq:wb_statistic}.
Hence, if experimenters want to control false positives, both neutral $\eta$ and \textbf{WB}-statistics can help. 

\paragraph{Apply \textbf{WB-TTTS} if regret of A/B testing is affordable}
It is worth to note that both \textbf{NB-TTTS} and \textbf{WB-TTTS} provide much smaller regret than uniform sampling. 
Therefore, regret with \textbf{WB-TTTS} and \textbf{NB-TTTS} are affordable, if experimenters can afford the much higher regret of A/B testing (uniform sampling). 
Table~\ref{tab:replay} presents performance on efficiency and trustworthiness with eBay real experiments replay (dataset $C$).
Though \textbf{TTTS} is not popular as \textbf{TS} in industry for adaptive traffic experimentation, it is actually a much promising alternative to reduce required sample size when experimenters can afford the regret in A/B testing. 
The power of identification under \textbf{WB-TTTS} and \textbf{NB-TTTS} are very close and much higher than others.
Compared with \textbf{NB-TTTS}, \textbf{WB-TTTS} has almost identical recall and provides better precision of identification in our real-world experiments. 
Hence, this paper promotes the novel \textbf{WB-TTTS} as better A/B testing alternative. 

\begin{table}[ht]
  \centering
  \begin{tabular}{|c|c|c|c|c|c|}
  \hline
                & FPR              & Power            & Precision & Regret           \\ \hline
  Uniform       & 12.8\% $\pm$ 2\% & 66.1\% $\pm$ 5\% & 83.8\%    & 85.7\% $\pm$ 0\% \\ \hline
  NB-TS         & 26.5\% $\pm$ 4\% & 87.5\% $\pm$ 5\% & 76.7\%    & 28.2\% $\pm$ 8\% \\ \hline
  WB-TS         & 10.4\% $\pm$ 3\% & 71.3\% $\pm$ 5\% & 87.3\%    & 31.8\% $\pm$ 8\% \\ \hline
  NB-TTTS       & 17.5\% $\pm$ 3\% & 97.6\% $\pm$ 5\% & 84.8\%    & 51.4\% $\pm$ 6\% \\ \hline
  WB-TTTS       & 12.6\% $\pm$ 3\% & 97.8\% $\pm$ 5\% & 88.6\%    & 53.5\% $\pm$ 6\% \\ \hline
  \end{tabular}
  \caption{Results from Real Experiments Replay ($\eta = 1$)}
  \label{tab:replay}
  \vspace{-12pt}
\end{table}

\begin{figure}[ht]
  \begin{subfigure}[b]{0.5\linewidth}
    \centering
    \includegraphics[width=\linewidth]{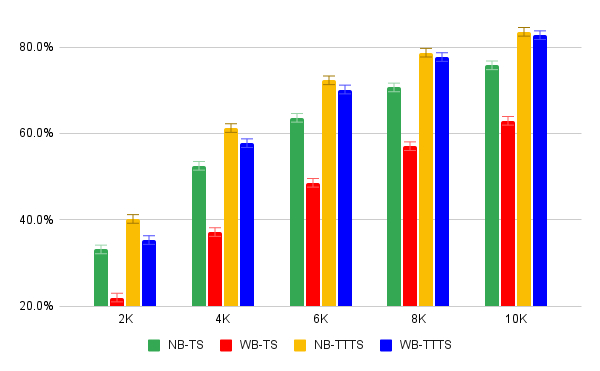} 
    \caption{Recall on Dataset $A$}
    \label{fig:stationary:a} 
  \end{subfigure}
  \begin{subfigure}[b]{0.5\linewidth}
    \centering
    \includegraphics[width=\linewidth]{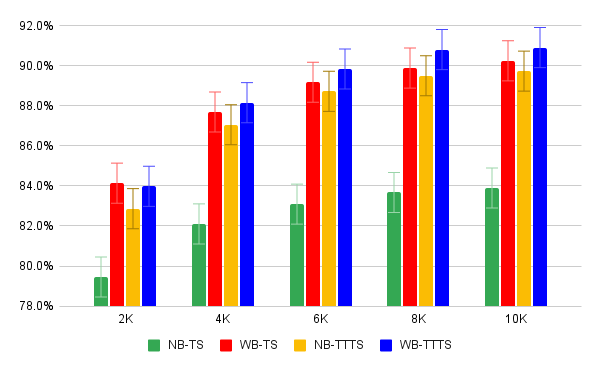} 
    \caption{Precision on Dataset $A$}
    \label{fig:stationary:b}
  \end{subfigure}
  \vskip\baselineskip
  \begin{subfigure}[b]{0.5\linewidth}
    \centering
    \includegraphics[width=\linewidth]{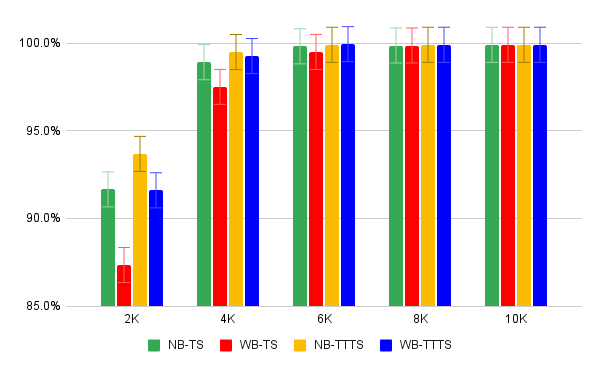} 
    \caption{Recall on Dataset $B$}
    \label{fig:stationary:c} 
  \end{subfigure}
  \begin{subfigure}[b]{0.5\linewidth}
    \centering
    \includegraphics[width=\linewidth]{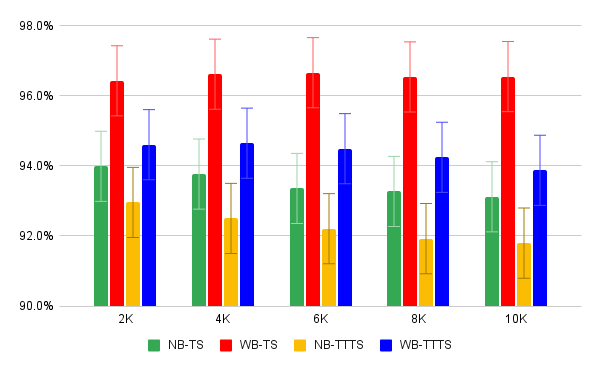} 
    \caption{Precision on Dataset $B$}
    \label{fig:stationary:d}
  \end{subfigure}
  \caption{
    Precision and Recall on Stationary Datasets
  }
  \Description[Stationary Case]{
    Stationary Case
  }
  \label{fig:stationary} 
  \vspace{-12pt}
\end{figure}

Further, Figure~\ref{fig:stationary} also shows that the promoted \textbf{WB-TTTS} has better precision with on-par recall.
The results on datasets $A$ and $B$ present two additional evidences when comparing the two \textbf{TTTS} sampling rules: 
(1) The proposed \textbf{WB-TTTS} shows higher precision than \textbf{NB-TTTS}, as \textbf{WB}-statistics also help control false positives; 
(2) The recall performance of \textbf{WB-TTTS} are almost on-par, when compared with \textbf{NB-TTTS}. 
The slight performance dropping of recall is probably because that biased $\hat{\theta}_{nb}(k)$ in \textbf{NB} can achieve the minimal variance than any weights of weighted batch statistics (Formula~\ref{eq:wb_statistic}).
Such findings suggest that \textbf{WB-TTTS} is practically more promising than \textbf{NB-TTTS}, since \textbf{WB-TTTS} can achieve much higher precision without significant loss on recall. 

\paragraph{Precisely control false positives may still be hard}
Although this paper presents that theoretically neutral posterior reshaping can help control false positive, the rule of thumbs $\eta=1,0.7$ seem to be not neutral enough to suppress false positives for \textbf{NB-TS} and \textbf{NB-TTTS} in real cases. 
Table~\ref{tab:fpr_inflation} shows that FPR may be still over the pre-determined $\rho=10\%$ with $\eta=1$ for eBay real experiments, while $\eta=0.7$ is too conservative and our system always considers $K^\prime = 2$ as it is the most common case. 
We believe that there exist two possible reasons, causing the failure:
(1) The number of equivalent best arms is unknown, while experimenters have to assume one $K^\prime$ as null hypothesis to apply neutral posterior reshaping $\eta$;
(2) The real world traffic is non-stationary on the trends, both \textbf{NB-TS} and \textbf{NB-TTTS} are probably not robust enough to hold neutral posterior reshaping $\eta$. 
In Table~\ref{tab:replay}, the two new sampling rules, \textbf{WB-TS} and \textbf{WB-TTTS}, control false positives better than \textbf{NB}-statistics based sampling rules in real world cases, which is probably due to robustness property of \textbf{WB}-statistics.
This reminds authors to compare the robustness of precision (related with FPR), recall and regret with non-stationary world. 

\begin{figure}[ht]
  \begin{subfigure}[b]{0.45\linewidth}
    \centering
    \includegraphics[width=\linewidth]{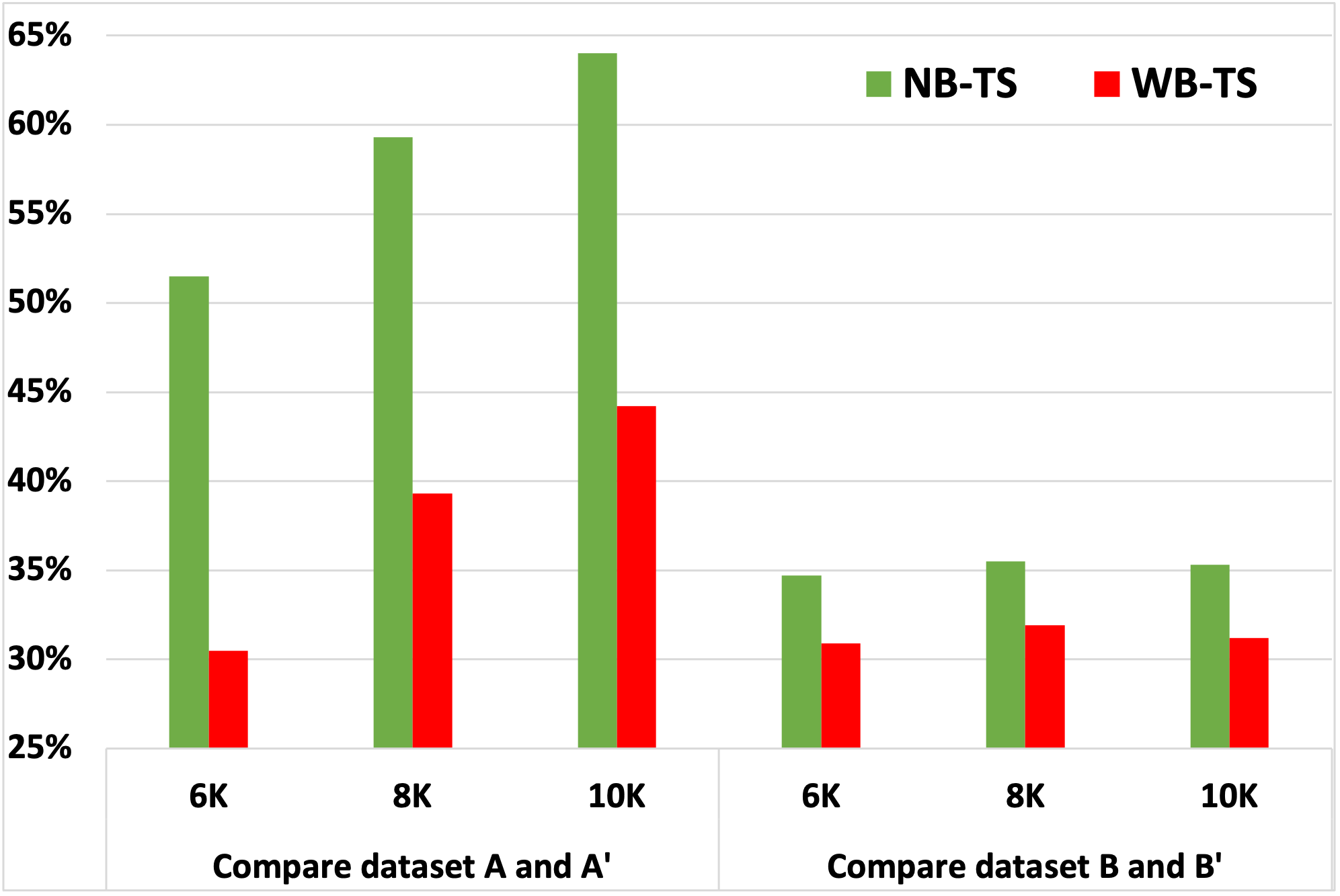} 
    \caption{Recall Changes (TS)}
    \label{fig:non-stationary:a} 
  \end{subfigure}
  \begin{subfigure}[b]{0.45\linewidth}
    \centering
    \includegraphics[width=\linewidth]{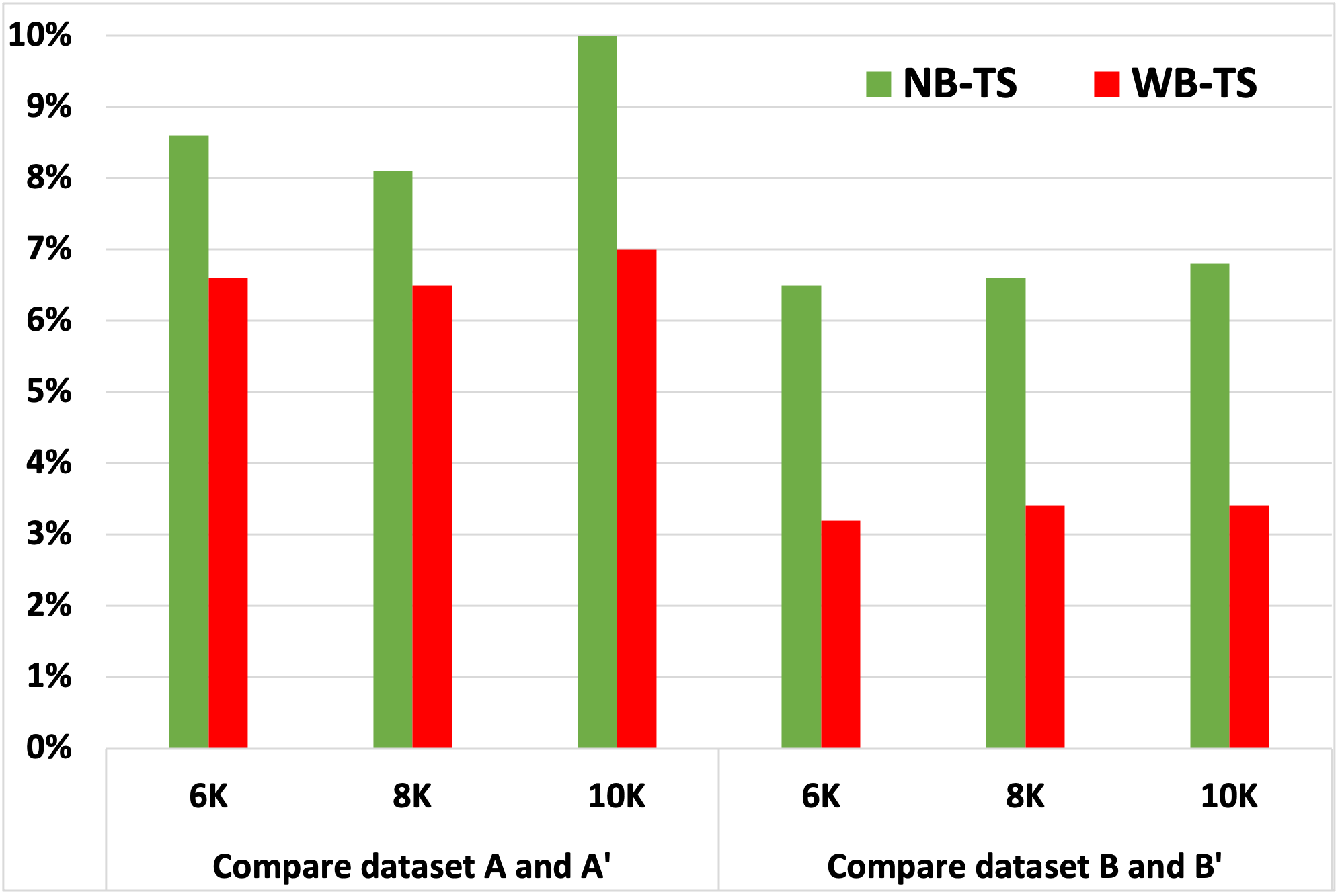} 
    \caption{Precision Changes (TS)}
    \label{fig:non-stationary:b} 
  \end{subfigure}
  \vskip\baselineskip
  \begin{subfigure}[b]{0.45\linewidth}
    \centering
    \includegraphics[width=\linewidth]{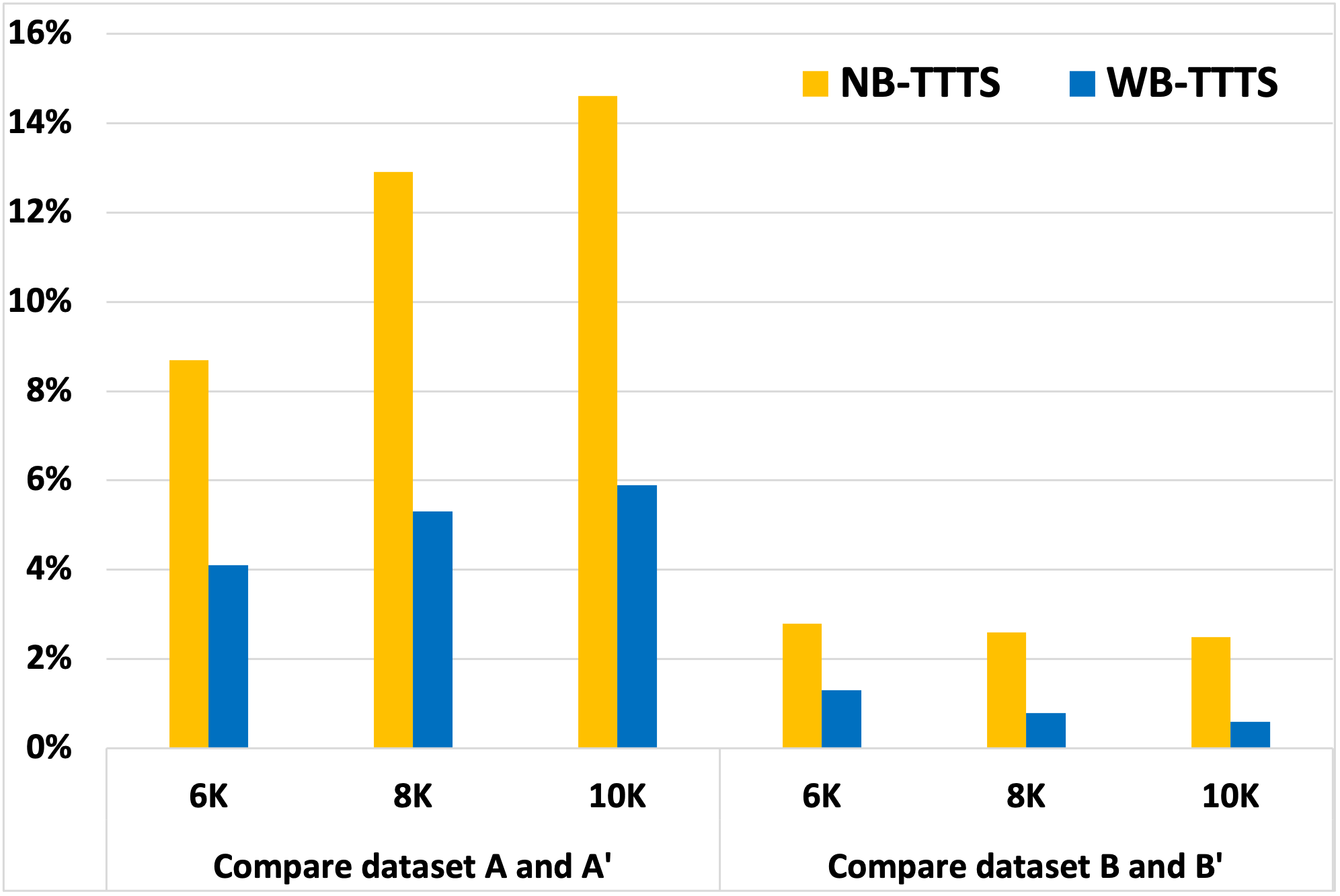} 
    \caption{Recall Changes (TTTS)}
    \label{fig:non-stationary:c}
  \end{subfigure}
  \begin{subfigure}[b]{0.45\linewidth}
    \centering
    \includegraphics[width=\linewidth]{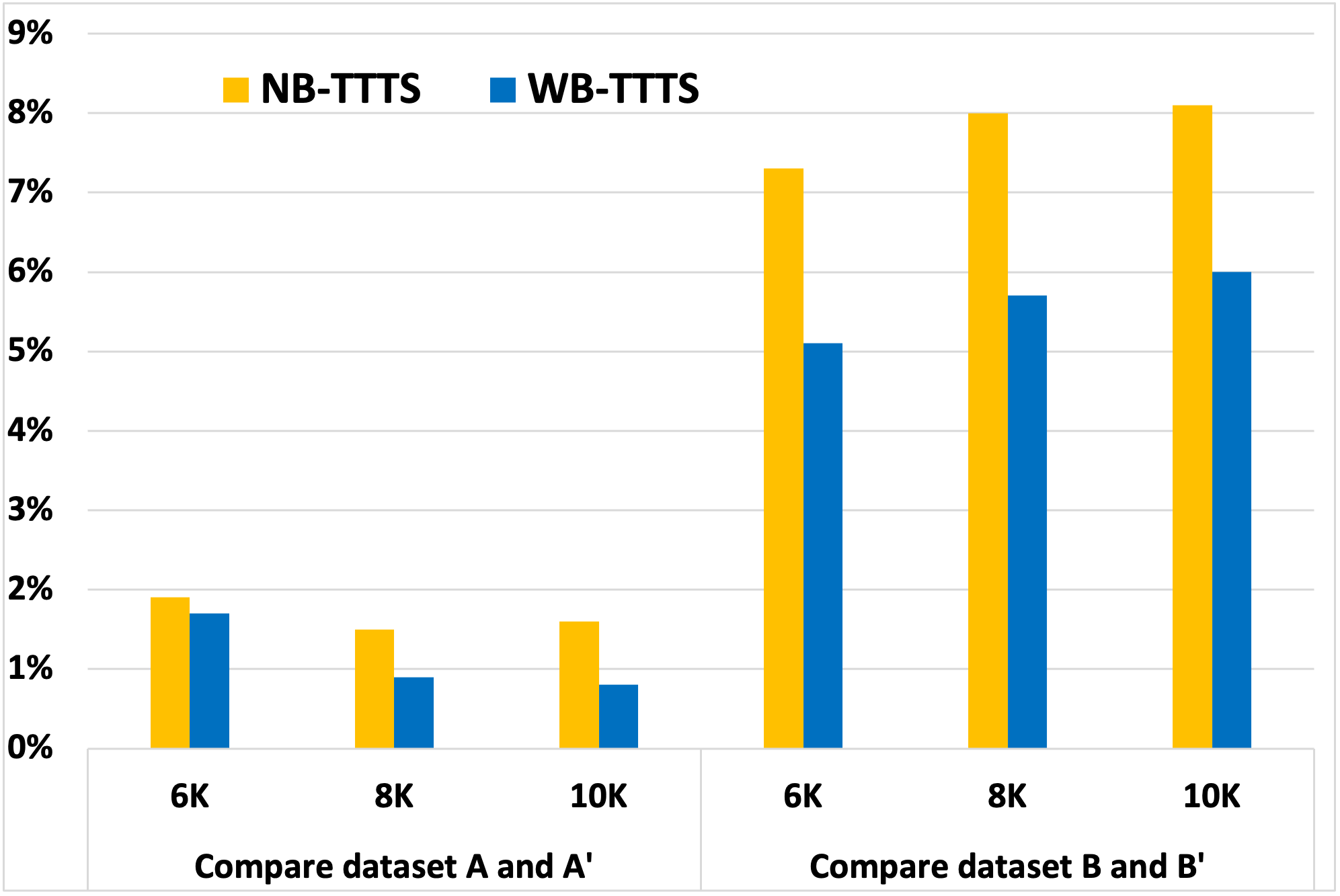} 
    \caption{Precision Changes (TTTS)}
    \label{fig:non-stationary:d}
  \end{subfigure}
  \caption{
    Performance Changes If Non-stationary
  }
  \Description[Non-Stationary Case]{
    Non-Stationary Cases
  }
  \label{fig:non-stationary}
  \vspace{-6pt}
\end{figure}

\paragraph{\textbf{WB}-statistics bring robustness against non-stationary trends}
The robustness can be measured as the changes of performance between stationary datasets and corresponding non-stationary datasets ($A$ and $A^{\prime}$, $B$ and $B^{\prime}$). 
If the sampling rule is robust, the dropping of performance shall be as small as possible. 
In Figure~\ref{fig:non-stationary}, we compare both precision and recall performance drop between the stationary and non-stationary world. 
It is quite obvious to conclude that \textbf{WB}-statistics based sampling rules, \textbf{WB-TTTS} and \textbf{WB-TS}, are more robust.
In Figure~\ref{fig:non-stationary:c}-\ref{fig:non-stationary:d}, \textbf{WB-TTTS} shows the most robust recall and precision, which can provide reliable or even better performance in the realistic non-stationary world. 

With non-stationary datasets, both \textbf{NB-TS} and \textbf{WB-TS} have loss of performance, especially on recall and regret. 
The loss of recall by \textbf{NB-TS} in Figure~\ref{fig:non-stationary:a} can be large (above $35\%$), which makes \textbf{NB-TS} eventually an even lower recall than \textbf{WB-TS}.
In Table~\ref{tab:regret}, \textbf{WB-TS} could even slightly beat \textbf{NB-TS} on the regret performance (34.4\% in $A^\prime$ and 18.9\% in $B^\prime$). 
The regret by \textbf{NB-TS} increases 10.8\% from 25.9\% ($A$) to 36.7\% ($A^\prime$) and increases 3.3\% from 16.1\% ($B$) to 19.4\% ($B^\prime$), while \textbf{WB-TS} only increases 6.4\% ($A^\prime$) and 2.0\% ($B^\prime$).

\begin{table}[]
  \centering
  \begin{tabular}{|c|c|c|c|c|c|}
  \hline
             & NB-TS             & WB-TS             & NB-TTTS           & WB-TTTS           \\ \hline
  $A$        & 25.9\% $\pm$ 8\%  & 28.1\% $\pm$ 8\%  & 47.5\% $\pm$ 8\%  & 49.9\% $\pm$ 5\%  \\ \hline
  $A^\prime$ & 36.7\% $\pm$ 17\% & 34.4\% $\pm$ 17\% & 53.9\% $\pm$ 4\%  & 54.2\% $\pm$ 4\%  \\ \hline
  $B$        & 16.1\% $\pm$ 5\%  & 16.9\% $\pm$ 5\%  & 27.7\% $\pm$ 12\% & 30.9\% $\pm$ 12\% \\ \hline
  $B^\prime$ & 19.4\% $\pm$ 10\% & 18.9\% $\pm$ 10\% & 35.3\% $\pm$ 12\% & 36.1\% $\pm$ 12\% \\ \hline
  \end{tabular}
  \caption{Regret on Stationary and Non-Stationary Datasets}
  \label{tab:regret}
  \vspace{-12pt}
\end{table}

\paragraph{Sample with \textbf{NB-TS} if minimizing regret is a must}
Both \textbf{NB-TS} and \textbf{WB-TS} are good choices when the cost of directing users to poor variations can be high, especially \textbf{NB-TS} in Table~\ref{tab:regret}. 
In Table~\ref{tab:replay}, regret performance with real experiments replays also suggest that \textbf{NB-TS} outperforms on minimizing regret trials. 
When experimenters have strong concerns of serving inferior arms for an experiment, \textbf{NB-TS} is the optimal choice to sample real traffic for data collection. 
Though sampling with \textbf{NB-TS}, we suggest computing \textbf{WB}-statistics for decision-making, as \textbf{WB}-statistics demonstrate higher precision and robustness property. 


\section{Conclusions}
\label{section:conclusions}
Everyone is eager for more efficient and more trustworthy experimentation to identify the best arm in industry.
As new web algorithms proliferate, experimentation platform faces an increasing demand on the velocity of online web experiments, 
which encourages eBay to efficiently allocate and manage traffic.
This paper shows that some adaptive traffic experimentation methodologies are promising alternatives to A/B testing, providing much more powerful performance when dealing with more variants. 
Especially, we proposed four Bayesian batch bandit algorithms (\textbf{NB-TS}, \textbf{WB-TS}, \textbf{NB-TTTS}, \textbf{WB-TTTS}) using summary batch statistics of one goal metric without incurring new engineering technical debts, and presented eBay learnings and novel findings, in terms of false positive rate, power, regret, and robustness.
Our results promote the novel algorithm \textbf{WB-TTTS} as the best A/B testing alternative among the candidates for delivering customer value as fast as possible.
The more important novel contribution of this paper is to bring trustworthiness of best arm identification algorithms into the quality evaluation criterion and highlight the existence of severe false positive inflation with equivalent best arms. 
This paper explicitly presents the theoretical analysis and simulation results on controlling false positives when there exist equivalent best arms, which are seldom discussed in literatures of either best arm identification or multi-armed bandit.
We also propose the novel neutral posterior reshaping to restore the symmetric flat Dirichlet distribution of convergence of posterior optimal probabilities, and to further control the false positives. 
Lastly, this paper presents lessons learned from eBay's experience, as well as comprehensive evaluations of the four Bayesian sampling algorithms, which cover trustworthiness, sensitivity, regret and robustness. 
We hope that this work will be useful to other industrial practitioners and inspire academic researchers interested in the trustworthiness of adaptive traffic experimentation.

\begin{acks}
The authors would like to gratefully acknowledge the leadership of the eBay Data Analytics Platforms organization for their continuous support in Experimentation and Multi-Armed Bandit projects. 
\end{acks}

\bibliographystyle{ACM-Reference-Format}
\bibliography{eval-bandit.bib}

\appendix

\section{Synthetic Experiments}
\label{appendix:synthetic}

\begin{table*}[ht]
  \centering
  \begin{tabular}{|c|l|l|}
  \hline
  Dataset & \multicolumn{1}{c|}{Type} & \multicolumn{1}{c|}{True Arm Means of an Experiment}                                                \\ \hline
          &                           & {[}0.872, 0.812, 0.433, 0.16, -0.125, -0.264, -0.306, -0.381, -0.536, -1.151{]}    \\ \cline{3-3} 
          &                           & {[}0.572, 0.451, 0.45, 0.291, 0.251, -0.061, -0.134, -0.342, -0.468, -0.55{]}      \\ \cline{3-3} 
          & $H_1$ is true             & {[}1.05, 0.846, 0.83, 0.371, 0.095, 0.025, -0.096, -0.318, -0.374, -0.444{]}       \\ \cline{3-3} 
          &                           & {[}0.626, 0.566, 0.466, 0.443, 0.256, 0.244, 0.143, -0.038, -0.149, -0.377{]}      \\ \cline{3-3} 
  A       &                           & {[}0.414, 0.381, 0.205, 0.115, 0.099, 0.093, 0.06, -0.1, -0.111, -0.153{]}         \\ \cline{2-3} 
          &                           & {[}0.731, 0.731, 0.567, 0.021, -0.086, -0.161, -0.192, -0.439, -0.55, -1.03{]}     \\ \cline{3-3} 
          &                           & {[}0.265, 0.265, 0.117, -0.006, -0.198, -0.336, -0.344, -0.346, -0.423, -0.559{]}  \\ \cline{3-3} 
          & $H_0$ is true             & {[}0.419, 0.419, 0.309, 0.293, 0.15, 0.06, -0.104, -0.175, -0.176, -0.571{]}       \\ \cline{3-3} 
          &                           & {[}1.093, 1.093, 0.76, 0.438, 0.158, 0.08, -0.252, -0.698, -0.722, -1.011{]}       \\ \cline{3-3} 
          &                           & {[}0.599, 0.599, 0.565, 0.212, 0.189, 0.093, 0.061, -0.188, -0.319, -0.335{]}      \\ \hline
          &                           & {[}0.82, 0.251, -0.028, -0.208, -0.421, -0.455, -0.529, -0.623, -0.897, -1.068{]}  \\ \cline{3-3} 
          &                           & {[}0.128, 0.005, -0.078, -0.118, -0.169, -0.319, -0.374, -0.439, -0.494, -0.594{]} \\ \cline{3-3} 
          & $H_1$ is true             & {[}0.866, 0.734, 0.386, 0.271, 0.251, 0.0, -0.157, -0.168, -0.422, -0.934{]}       \\ \cline{3-3} 
          &                           & {[}0.639, 0.254, 0.217, 0.163, 0.108, -0.02, -0.066, -0.21, -0.317, -0.929{]}      \\ \cline{3-3} 
  B       &                           & {[}0.792, 0.348, 0.033, -0.039, -0.046, -0.095, -0.191, -0.549, -1.017, -1.33{]}   \\ \cline{2-3} 
          &                           & {[}1.146, 1.146, 1.146, 0.588, 0.276, 0.27, 0.021, -0.01, -0.298, -0.559{]}        \\ \cline{3-3} 
          &                           & {[}1.116, 1.116, 1.116, 0.68, 0.185, 0.056, -0.077, -0.135, -0.711, -1.217{]}      \\ \cline{3-3} 
          & $H_0$ is true             & {[}0.5, 0.5, 0.5, 0.306, 0.044, 0.024, -0.037, -0.188, -0.191, -0.415{]}           \\ \cline{3-3} 
          &                           & {[}0.421, 0.421, 0.421, 0.368, 0.262, 0.023, -0.327, -0.339, -0.72, -1.02{]}       \\ \cline{3-3} 
          &                           & {[}0.684, 0.684, 0.684, 0.624, 0.609, 0.412, 0.175, -0.202, -0.231, -0.692{]}      \\ \hline
  \end{tabular}
  \caption{Synthetic Arms In Dataset $A$ and $B$}
  \label{tab:synthetic-arms}
  \vspace{-12pt}
\end{table*}

Both synthetic datasets $A$ and $B$ contain ten experiments with five experiments with a single best arm, where the true arm means are drawn randomly from $\mathcal{N}(0,0.5)$. 
For the rest of the experiments, there exist five experiments with equivalent $K^{\prime}=2$ best arms in dataset $A$ and equivalent $K^{\prime}=3$ best arms in dataset $B$. 

To generate an experiment with equivalent best arms, we firstly randomly sample from $\mathcal{N}(0,0.5)$ as arm means of 10 arms.
Then, for each experiment, we replace the second-best arm with the best arm for dataset $A$, and replace both the second and third best arms for dataset $B$.
The generated dataset $A$ and $B$ are shown in Table~\ref{tab:synthetic-arms}.

For non-stationary dataset $A^{\prime}$ and $B^{\prime}$, same arm means in the corresponding dataset ($A$ or $B$) are used as initial values.
Then, arm means $\theta^*$ per experiment over batch index $b$ are by formula 
\begin{equation}
  \begin{aligned}
    \label{eq:arm_means}
    \theta^* + 0.5[\cos(\frac{\pi}{20}(b-1)) - 1]
  \end{aligned}
\end{equation}
where arm means monotonically decreases within total 20 batches. 

\begin{figure}[ht]
  \centering
  \includegraphics[width=0.9\linewidth]{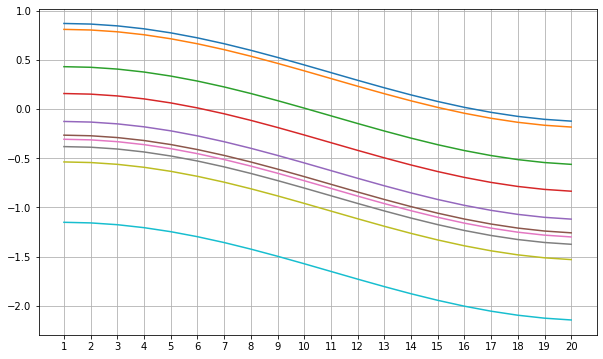}  
  \caption{
    An Experiment of Non-Stationary Means in $A^{\prime}$
  }
  \Description[Non-Stationary Reward Trend]{
    Non-Stationary Reward Trend
  }
  \label{fig:reward-trend}
  \vspace{-12pt}
\end{figure}

Arms in datasets $A^{\prime}$ and $B^{\prime}$ are non-stationary over time but always with the same trending. The difference in true expected rewards between any of the two arms does not change (see Figure~\ref{fig:reward-trend} as an experiment in dataset $A^{\prime}$). 

\section{Naive Batch Bayesian Updating}
\label{appendix:naive_batch}
Under the classical CLT, the asymptotical normality of batch statistics can be written as 
$\sqrt{n_{b,k}}(\bar{Y}_{b,k} - \theta^*_{k}) \xrightarrow{d} \mathcal{N}(0,\sigma_{k}^{2})$ with known variance $\sigma_{k}^{2}$.

Hence, an asymptotical and convenient Gaussian likelihood can be derived based on above asymptotical normality of batch mean,
\begin{equation}
  \label{eq:likelihood_of_batch}
  \begin{aligned}
    p(\bar{Y}_{b,k} \mid \theta_{k}; \sigma_{k}, n_{b,k}) = \frac{\sqrt{n_{b,k}}}{\sigma_{k}\sqrt{2\pi}} \exp[-\frac{1}{2} \frac{n_{b,k}}{\sigma^2_{k}} (\bar{Y}_{b,k} - \theta_{k})^2]
  \end{aligned}
\end{equation}
where the only parameter is $\theta_{k}$ and a conjugate gaussian prior $\pi_0(\theta_{k}) = \mathcal{N}(\mu_{0,k}, \frac{1}{\tau_{0,k}})$ can be used for convenience with two hyperparameters $\mu_{0,k}$ and $\tau_{0,k}$ to be updated. 
It is proportional to the likelihood used in Formula \ref{eq:naive_iterative_update} with a simple transformation.

With the collected one batch of data ${n_{b,k}, \bar{Y}_{b,k}}$, we can derive the posterior $\pi_b(\theta_k)$ from the prior $\pi_{b-1}(\theta_k) = \mathcal{N}(\mu_{b-1,k}, \frac{1}{\tau_{b-1,k}})$,
\begin{equation}
  \label{eq:naive_update_1}
  \begin{aligned}
    \pi_b(\theta_k) &\propto \pi_{b-1}(\theta_k) p(\bar{Y}_{b,k} \mid \theta_{k}; \sigma_{k}, n_{b,k}) \\
    &\propto \exp[-\frac{1}{2} ( \frac{n_{b,k}}{\sigma^2_{k}} (\bar{Y}_{b,k} - \theta_{k})^2 + \tau_{b-1,k}(\theta_{k} - \mu_{b-1,k})^2)] \\
    &\approx \exp[-\frac{1}{2}\tau_{b,k}(\theta_{k} - \mu_{b,k})^2] \\
    &\propto \mathcal{N}(\mu_{b,k}, \frac{1}{\tau_{b,k}})
  \end{aligned}
\end{equation}
By matching the coefficients, we derived the below iterated updating formula, 
\begin{equation}
\label{eq:naive_iterate_1}
\begin{aligned}
  \tau_{b,k} &= \tau_{b-1,k} + \frac{1}{\sigma^2_{k}} n_{b,k} \\
  \mu_{b,k} &= \frac{\mu_{b-1,k}\tau_{b-1,k} + \frac{1}{\sigma^2_{k}} n_{b,k}\bar{Y}_{b,k}}{\tau_{b-1,k} + \frac{1}{\sigma^2_{k}} n_{b,k}}
\end{aligned}
\end{equation}

Further, we can derive the cumulative Bayesian updating Formula~\ref{eq:naive_cumulative_update} for posterior distribution $\pi_b(\theta_k) = \mathcal{N}(\mu_{b,k}, \frac{1}{\tau_{b,k}})$.
\begin{equation}
  \label{eq:naive_cumulative_update}
  \begin{aligned}
    \tau_{b,k} &= \tau_{0,k} + \frac{1}{\sigma^2_{k}} \sum_{i=1}^{b}{n_{i,k}} \\
    \mu_{b,k} &= \frac{\mu_{0,k}\tau_{0,k} + \frac{1}{\sigma^2_{k}} \sum_{i=1}^{b}{n_{i,k}\bar{Y}_{i,k}}}{\tau_{0,k} + \frac{1}{\sigma^2_{k}}\sum_{i=1}^{b}{n_{i,k}}}
  \end{aligned}
\end{equation}
For convenience, we name Algorithm~\ref{algo:batch_bandit} with Bayesian updating by Formula~\ref{eq:naive_cumulative_update} as \textbf{NB}.

\section{Proper Weighting of Batches}
\label{appendix:wb_statistics}
In this section, we want to show that $\phi_{i,k}=1$ and $\phi_{i,k}=\sqrt{T_i}$ are both proper weighting for Weighted Batch statistics by Formula~\ref{eq:wb_statistic}.

Vitor Hadad et al. \cite{hadad2021confidence} derived a class of estimators with asymptotical normality property.
Weighted Batch statistics belongs to such class of estimators (Formula 25 of Vitor Hadad et al. \cite{hadad2021confidence}),
\begin{equation}
  \label{eq:wb_sample_statistic}
  \begin{aligned}
    \hat{\theta}_k &= \sum_{b=1}^{B}{w_{b,k} \bar{Y}_{b,k}} = \frac{\sum_{b=1}^{B}{\phi_{b,k} \sqrt{n_{b,k}} \bar{Y}_{b,k}}}{\sum_{b=1}^{B}{\phi_{b,k} \sqrt{n_{b,k}}}}\\
    &= \frac{ \sum_{b=1}^{B} \sum_{i=1}^{T_b} \frac{\phi_{b,k}}{\sqrt{n_{b,k}}} \mathbb{I}[I_{b,i}==k]Y_{b,i} }{ \sum_{b=1}^{B} \sum_{i=1}^{T_b} \frac{\phi_{b,k}}{\sqrt{n_{b,k}}} \mathbb{I}[I_{b,i}==k] } \\
    &= \frac{ \sum_{b=1}^{B} \sum_{i=1}^{T_b} \frac{h_{b,k}}{e_{b,k}} \mathbb{I}[I_{b,i}==k]Y_{b,i} }{ \sum_{b=1}^{B} \sum_{i=1}^{T_b} \frac{h_{b,k}}{e_{b,k}} \mathbb{I}[I_{b,i}==k] }
  \end{aligned}
\end{equation}
where $h_{b,k}$ is the variance stabilizing weight proposed by Vitor Hadad et al. \cite{hadad2021confidence}.
And, the relationship between $h_{b,k}$ and $\phi_{b,k}$ becomes $\frac{h_{b,k}}{e_{b,k}} = \frac{\phi_{b,k}}{\sqrt{n_{b,k}}}$ for batch $b$ and arm $k$.

According to the Theorem 2 from Vitor Hadad et al. \cite{hadad2021confidence}, variance stabilizing weight $h_{b,k}$ or batch weight $\phi_{b,k}$ must satisfy the variance convergence assumption to ensure the asymptotical normality property. 
By re-formulating the original assumption with the fact that $n_{b,k} = T_b e_{b,k}$ under batch settings, our batch weight $\phi_{b,k}$ is required to satisfy below assumption
\begin{equation}
  \label{eq:variance_convergence}
  \begin{aligned}
    \frac{ \sum_{b=1}^{B} \sum_{i=1}^{T_b} \frac{\phi_{b,k}^2}{T_b} }{ \mathbb{E}[\sum_{b=1}^{B} \sum_{i=1}^{T_b} \frac{\phi_{b,k}^2}{T_b}] } 
    \xrightarrow{T \rightarrow \infty} 1
  \end{aligned}
\end{equation}
where $T=\sum_{b=1}^{B}{T_b}$ represents the total cumulative sample size.

With the choice of $\phi_{b,k}=1$, it is easy to see that the numerator of Assumption~\ref{eq:variance_convergence} is constant as below
\begin{equation}
  \begin{aligned}
    \label{eq:variance_convergence_1}
    \sum_{b=1}^{B} \sum_{b=1}^{T_b} \frac{\phi_{b,k}^2}{T_b} = B
  \end{aligned}
\end{equation}
hence, $\phi_{b,k}=1$ can satisfy variance convergence assumption under any batch settings.
So, $\phi_{b,k}=1$ is one of the proper batch weighting with asymptotically normality property.

With the choice of $\phi_{b,k}=\sqrt{T_b}$, the numerator of Assumption~\ref{eq:variance_convergence} becomes the following
\begin{equation}
  \begin{aligned}
    \label{eq:variance_convergence_2}
    \sum_{b=1}^{B} \sum_{i=1}^{T_b} \frac{\phi_{b,k}^2}{T_b} = \sum_{b=1}^{B}{T_b}
  \end{aligned}
\end{equation}

Under the fixed sample size batch settings, $T_b=\lambda$ is a constant with $\lambda$ sample size per batch.
Hence, Formula~\ref{eq:variance_convergence_2} is a constant and $\phi_{b,k}=\sqrt{T_b}$ can satisfy variance convergence assumption under fixed sample size settings.

Under the fixed duration batch settings, we have $T_b \sim Pois(\lambda)$ where each batch traffic size is independent.
Then, the summation of Poisson-distributed random variables by Formula~\ref{eq:variance_convergence_2} will be,
\begin{equation}
  \begin{aligned}
    \label{eq:variance_convergence_3}
    \sum_{b=1}^{B}{T_b} \sim Pois(B\lambda)
  \end{aligned}
\end{equation}
where $\mathbf{E}[\sum_{b=1}^{B}{T_b}] = \mathbf{Var}[\sum_{b=1}^{B}{T_b}] = B\lambda$. 
Then, the new random variable $\frac{\sum_{b=1}^{B}{T_b}}{B\lambda}$ has $\mathbf{E}[\frac{\sum_{b=1}^{B}{T_b}}{B\lambda}]=1$ and $\mathbf{Var}[\frac{\sum_{b=1}^{B}{T_b}}{B\lambda}]=\frac{1}{B\lambda}$.
Hence, we can have $\frac{\sum_{b=1}^{B}{T_b}}{B\lambda} \rightarrow 1$ to satisfy the variance convergence assumption.
So, $\phi_{b,k}=\sqrt{T_b}$ is also a proper batch weighting.

\section{Quality of Identification Decision}
\label{appendix:quality}
The testing procedure can be seen as making decision among multiple hypothesis as below.
\begin{equation}
  \label{eq:hypothesis}
  \begin{aligned}
    H_0 &\colon \text{$K^{\prime}$ arms are equivalent best} \\
    H_k &\colon I_{1}^{*} = k
  \end{aligned}
\end{equation}
where $H_0$ is the null hypothesis and there exist total $K$ alternative hypothesis.
And, each alternative hypothesis $H_k$ states that the arm $k$ is the best. 
Without loss of generality, we let arm $k=1$ as the best arm to be identified ($H_1$ is true) for situation that $H_0$ is false and true means of arms are in the decreasing order $\theta_1^* \ge \theta_2^* \ge \dots \ge \theta_K^*$.

\paragraph{False Positive Rate}
FPR (False Positive Rate) is the probability of falsely rejecting the null hypothesis $H_0$.
With $B$ batches of data per test, FPR is a ratio as in Formula~\ref{eq:fpr} calculated when $H_0$ is true, 
\begin{equation}
  \label{eq:fpr}
  \begin{aligned}
    FPR = \frac{\sum_{i}^{N_{0}}{\mathbb{I}[\max_{1 \le k \le K}{\alpha_{k, i}} > \delta]}}{N_{0}}
  \end{aligned}
\end{equation}
where $N_0$ is the number of tests that $H_0$ is true.

\paragraph{Power of Identification}
With $B$ batches of trials, Formula~\ref{eq:power} defines Power (a.k.a. Recall) as the probability of correctly identifying the best arm in a test where $H_1$ is true.
\begin{equation}
  \label{eq:power}
  \begin{aligned}
    Power = \frac{\sum_{i}^{N_{1}}{\mathbb{I}[\alpha_{1, i} > \delta]}}{N_{1}}
  \end{aligned}
\end{equation}
where $N_1$ is the number of tests that $H_1$ is true.

\paragraph{Precision of Identification}
Precision is related to the false positives, but also depends upon the distribution of $H_0$ and $H_1$.
Formula~\ref{eq:precision} defines precision as the fraction of correct best arm identifications among total best arm claims.
\begin{equation}
  \label{eq:precision}
  \begin{aligned}
    Precision = \frac{ \sum_{i}^{N_{1}}{\mathbb{I}[\alpha_{1, i} > \delta]} }{ \sum_{i}^{N_{0}+N_{1}}{\mathbb{I}[\max_{1 \le k \le K}{\alpha_{k, i}} > \delta]} }
  \end{aligned}
\end{equation}

\paragraph{Regret}
Regret trials are samples exposed to inferior arms during a test. 
For a test that $H_1$ is true, regret Formula~\ref{eq:regret} is the proportion of cumulative regret trials over the exposed total $B$ batches of trials.
\begin{equation}
  \label{eq:regret}
  \begin{aligned}
    Regret = \frac{\sum_{b=1}^{B} \sum_{i}^{T_b} {\mathbb{I}[I_{b, i} \ne 1]}}{\sum_{b=1}^{B}{T_b}}
  \end{aligned}
\end{equation}
By taking average over $N_1$ tests that $H_1$ is true, we can report the expected regret $\mathbb{E}[Regret]$ as our regret measurement.

\end{document}